\documentclass[runningheads]{llncs}

\usepackage{eccv}

\usepackage{eccvabbrv}

\usepackage{graphicx}
\usepackage{booktabs}

\usepackage[accsupp]{axessibility}  

\usepackage{cite}
\usepackage{wrapfig}
\usepackage{overpic}
\usepackage{graphicx}
\usepackage{amsmath}
\usepackage{amssymb}
\usepackage{pifont}
\usepackage{bm}
\usepackage{subcaption}
\usepackage{multirow}
\usepackage{tcolorbox}
\usepackage[font=small,labelfont=bf]{caption}
\usepackage{algorithmic}
\usepackage{algorithm}
\usepackage{caption}

\newcommand{\Paragraph}[1]{\vspace{0.6mm} \noindent \textbf{#1} \hspace{0mm}}
\newcommand{\Section}[1]{\vspace{-1mm} \section{#1} \vspace{-1mm}}
\newcommand{\SubSection}[1]{\vspace{-1mm} \subsection{#1} \vspace{-1mm}}

\newcommand{\multiline}[1]{%
    \begin{tabularx}{\dimexpr\linewidth-\ALG@thistlm}[t]{@{}X@{}}
        #1
    \end{tabularx}
}

\usepackage{xcolor}
\definecolor{citecolor}{RGB}{34,139,34}
\definecolor{lightred}{RGB}{255,100,100}
\definecolor{cell_bisque}{rgb}{1.0, 0.89, 0.77}
\definecolor{cell_blond}{rgb}{0.98, 0.94, 0.75}
\definecolor{cell_blue}{RGB}{155, 187, 228}
\definecolor{princetonorange}{rgb}{1.0, 0.56, 0.0}
\definecolor{pinkpearl}{rgb}{0.91, 0.67, 0.81}
\definecolor{mossgreen}{rgb}{0.68, 0.87, 0.68}

\usepackage[pagebackref,breaklinks,colorlinks,citecolor=eccvblue]{hyperref}

\usepackage{orcidlink}

\begin{document}

\title{Dense-Face: Personalized Face Generation Model via Dense Annotation Prediction} 

\titlerunning{Personalized Face Generation Model via Dense Annotation Prediction}

\author{Xiao Guo\inst{1}\orcidlink{0000-0003-3575-3953} \and
Manh Tran\inst{1} \and
Jiaxin Cheng\inst{2} \and
Xiaoming Liu\inst{1} \orcidlink{0000-0003-3215-8753}
}

\authorrunning{X. Guo et al.}

\institute{Michigan State University \and
University of Southern California, Information Sciences Institute
\email{\{guoxia11,tranman1,liuxm\}@msu.edu}, \email{chengjia@isi.edu}\\
\href{https://chelsea234.github.io/Dense-Face.github.io/}{Dense-Face.github.io}
}
\maketitle

\noindent\begin{minipage}{\linewidth}
  \centering
  \begin{overpic}[width=1\textwidth]{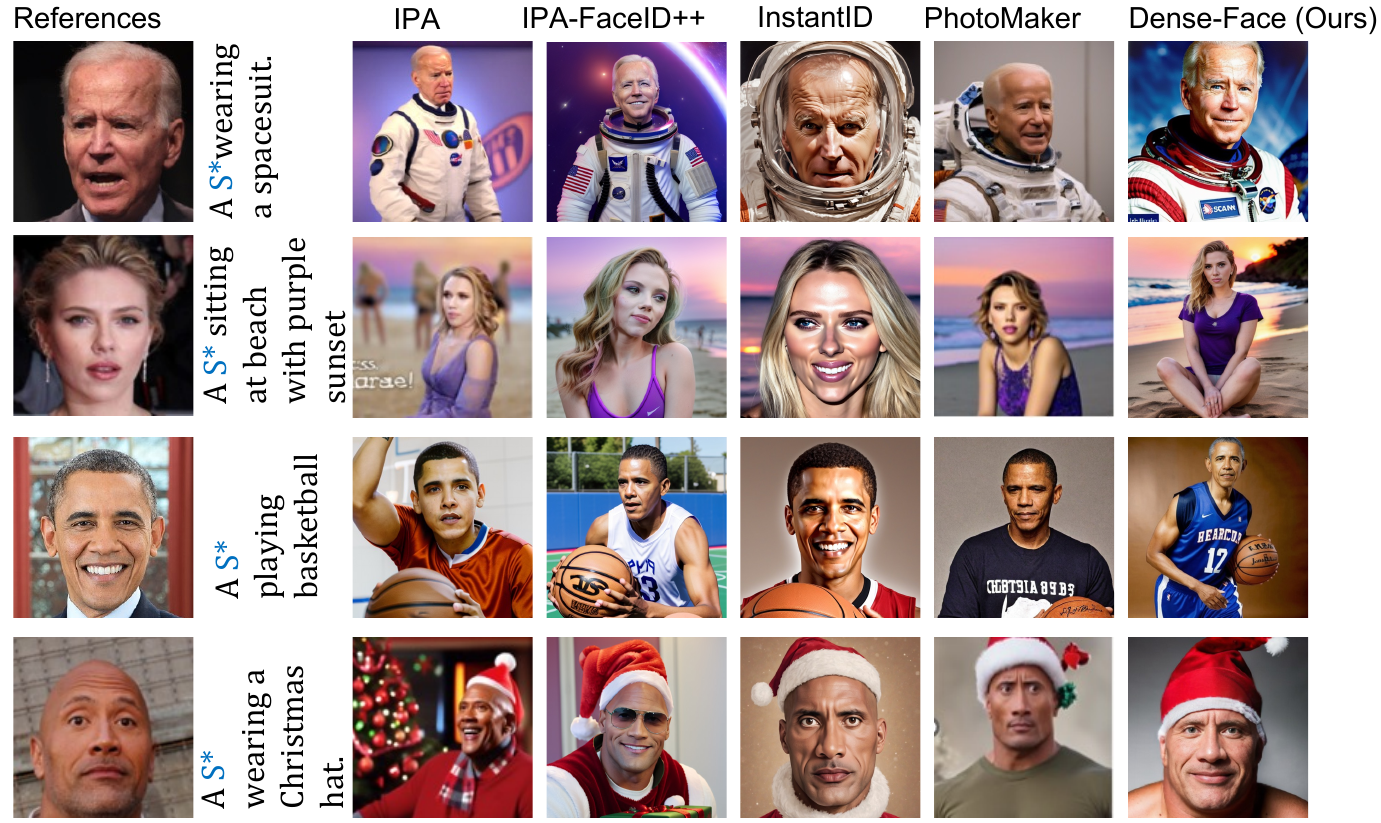}
    \put(32,57.5){\tiny{\cite{ye2023ip-adapter}}}
    \put(50.5,57.5){\tiny{\cite{ye2023ip-adapter}}}
    \put(62.5,57.5){\tiny{\cite{wang2024instantid}}}
    \put(78,57.5){\tiny{\cite{li2023photomaker}}}
  \end{overpic}
    \captionof{figure}{Performance comparison of different personalized methods.}
  \label{fig:show_case_1}
\end{minipage}
\begin{abstract}
    The text-to-image (T2I) personalization diffusion model can generate images of the novel concept based on the user input text caption.
    However, existing T2I personalized methods either require test-time fine-tuning or fail to generate images that align well with the given text caption.
    In this work, we propose a new T2I personalization diffusion model, Dense-Face, which can generate face images with a consistent identity as the given reference subject and align well with the text caption. 
    Specifically, we introduce a pose-controllable adapter for the high-fidelity image generation while maintaining the text-based editing ability of the pre-trained stable diffusion (SD). Additionally, we use internal features of the SD UNet to predict dense face annotations, enabling the proposed method to gain domain knowledge in face generation. 
    Empirically, our method achieves state-of-the-art or competitive generation performance in image-text alignment, identity preservation, and pose control.
\end{abstract}
\Section{Introduction}
The Text-to-image (T2I) personalized generation, or customized T2I diffusion models, has obtained significant attention for its ability to generate visually compelling images for novel concepts unseen in training~\cite{gal2022image_textual_inversion,dreambooth,custom_diffusion,Gligen,controlNet}. 
One of the essential personalized T2I generation applications is face generation~\cite{karras2020analyzing,karras2019style,chan2022efficient,choi2020starganv2,li2019faceshifter}, a classic and challenging computer vision task, requiring the generative method to both model complicated non-rigid face characteristics (\textit{e.g.}, subject identity) and produce high-quality images. 

Prior research works have successfully adapted the Stable Diffusion (SD) model for face generation, producing high-fidelity and identity-preserved images with high text-controllability. First, test-time fine-tuning (TTFT) is common to model the subject appearance~\cite{custom_diffusion,gal2022image_textual_inversion,custom_diffusion,gal2023encoder}, generating realistic customized images in various contexts. 
Based upon that, recent works~\cite{gal2022image_textual_inversion,dreambooth,custom_diffusion,controlNet,wang2024stableidentity} introduce a new identity preservation face generation scheme without TTFT. 
Specifically, they first integrate the deterministic face identity embedding with the text-embedding input to the pre-trained SD.
Then, they fine-tune the pre-trained SD on face samples from LAION$5$B~\cite{schuhmann2022laion} or other complicated collection procedures, for learning the face generation domain knowledge.
This work focuses on the second generation paradigm, since multiple face images of the test subject can often be unavailable for TTFT. 
Specifically, we empirically notice the previous work is limited in achieving optimal performance on both face domain generation and text-based editing. 
This is because they uniformly fine-tune the pre-trained SD on the face-specific dataset. 
This fine-tuning step may decrease the text-editing ability of the pre-trained SD, and the resultant model's text controllability depends on dataset qualities that vary from one to another. 

\begin{figure*}[t]
  \centering
  \begin{subfigure}[b]{1.\linewidth}
  \includegraphics[width=\linewidth]{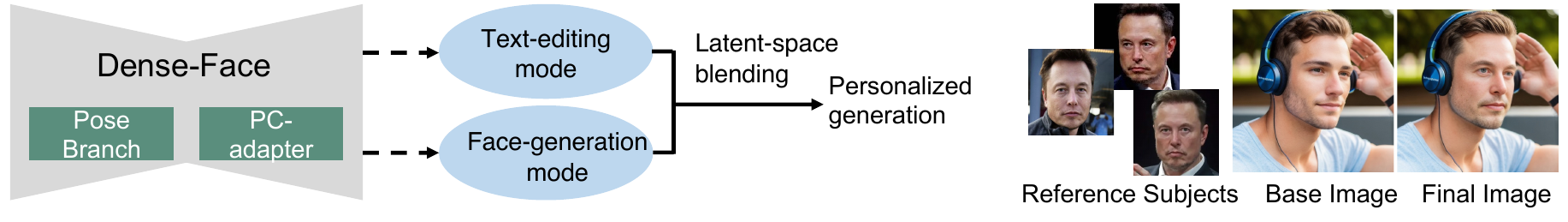}
  \caption{}
  \label{fig:overview_balance}
  \end{subfigure}
  \begin{subfigure}[b]{0.95\linewidth}
  \includegraphics[width=1\linewidth]{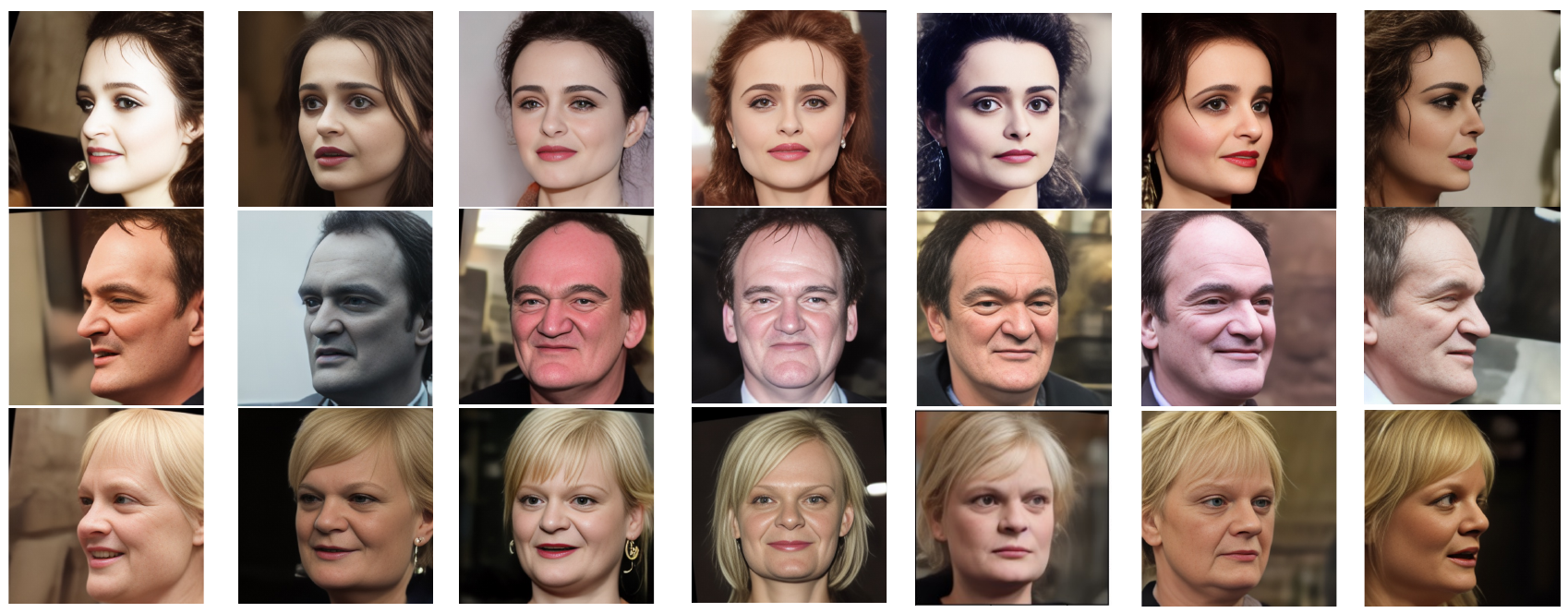}
  \caption{\vspace{-3mm}}
  \label{fig:pose_controllable}
  \end{subfigure}
  \caption{
  (a) Our proposed Dense-Face introduces additional components, including a pose branch and PC-adapter, on the top of the pre-trained SD. These two components enable Dense-Face to have two generation modes: \textit{text-editing mode} and \textit{face-generation mode}. These two modes are jointly used via the latent space blending (Sec.~\ref{sec:inference}) for the personalized generation. For example, given one of reference subject images, the text-editing mode generates a base image, and face-generation mode updates the face region for the identity-preservation in the final result.
  (b) Dense-Face in face-generation mode generates realistic face images at different pose views. 
  \vspace{-7mm}}
\end{figure*}

To address this limitation, we propose Dense-Face, a new personalization model, leveraging a novel pose-controllable adapter (PC-adapter) and the additional pose branch to help the proposed method obtain two generation modes: \textit{text-editing mode} and \textit{face-generation mode} (as depicted in Fig.~\ref{fig:overview_balance}). 
Then, these two generation modes are jointly used via a latent space blending technique for the personalized image generation, which achieves the domain-specific generation while maintaining the premium pre-trained SD's text controllability. Moreover, we utilize the internal SD-UNet feature for predicting dense face annotations, enhancing the face generation domain knowledge further. 

\begin{figure*}[t]
  \centering
  \includegraphics[width=\linewidth]{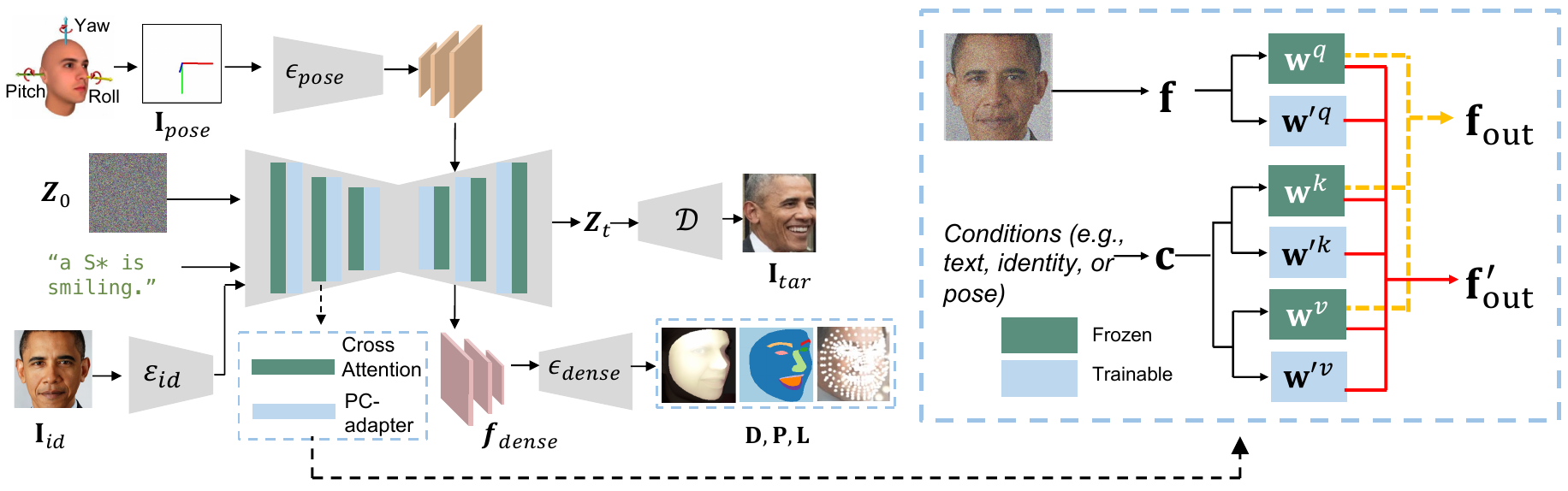}
  \caption{
      We propose Dense-Face for personalized image generation, which introduces additional components, such as a pose-controllable (PC) adapter, pose branch (\textit{i.e.}, $\bm{\epsilon}_{pose}$) and annotation prediction module (\textit{i.e.}, $\bm{\epsilon}_{dense}$) on the top of the pre-trained T2I-SD.
      The input includes captions, head pose and reference image \textit{(i.e.}, $\mathbf{I}_{pose}$ and $\mathbf{I}_{id}$). The output includes generated faces ($\mathbf{I}_{tar.}$) and dense face annotations (\textit{e.g.}, face depths ($\mathbf{D}$), pseudo masks ($\mathbf{P}$), and landmarks ($\mathbf{L}$)). 
      We only train $\bm{\epsilon}_{pose}$, $\bm{\epsilon}_{dense}$ and the PC adapter in training and freeze the pre-trained SD.
      The PC-adapter (${\mathbf{w}^{\prime}}^{q}, {\mathbf{w}^{\prime}}^{v}$, and ${\mathbf{w}^{\prime}}^{v}$) modifies cross attention module ($\mathbf{w}^{q}, \mathbf{w}^{v}$, and $\mathbf{w}^{k}$)
      forward propagation, from $\mathbf{f}_{out}$ (\textcolor{orange}{orange dash line}) to $\mathbf{f}^{\prime}_{out}$ (\textcolor{red}{red solid line}).
      $\bm{\epsilon}_{dense}$ utilizes the internal UNet features (\textit{i.e.}, $\mathbf{f}_{dense}$) for predicting dense face annotations. 
      \vspace{-4mm}
    }
    \label{fig:overview_architecture}
\end{figure*}

Specifically, our proposed Dense-Face, as depicted in Fig.~\ref{fig:overview_architecture}, introduces a pose branch, the PC-adapter, and the annotation prediction module on the top of the pre-trained SD. The PC-adapter and pose branch alter the forward propagation of cross-attention modules, adapting the pre-trained SD for the face generation domain. We keep the pre-trained SD frozen and only update additional components, which are plug-in modules that enable Dense-Face to have two generation modes:\textit{ text-editing mode} and \textit{face-generation mode}. The former mode generates the base image with diverse contexts based on the text caption; then the latter generation mode updates the face region of the base image to preserve the subject identity. Such a two-stage design relies on the generation ability of face-generation mode in generating faces at different poses given the head pose (\textit{i.e.}, Euler angles consisting of yaw, pitch and roll) as the condition (Fig.~\ref{fig:pose_controllable}).
Also, this procedure involves a commonly used latent space blending technique~\cite{avrahami2022blended,avrahami2023blended,liu2024towards,zhang2023paste} that effectively preserves the background contexts (Sec.~\ref{sec:inference}).
In this way, our Dense-Face improves the specific face domain generation while maintaining the original T2I-SD text-based editing ability. 

Secondly, given the sparse head pose condition (\textit{i.e.}, Euler angles), Dense-Face employs the annotation prediction module that leverages the internal SD feature for predicting different dense face annotations to improve further the face generation domain knowledge (Sec.~\ref{sec:method_dense}). This is inspired by the fact that regularizing such internal SD features assists the domain-specific generation in prior works~\cite{gu2023photoswap,peng2023portraitbooth,kim2023dense}. These features contain rich and detailed object semantics (including human face) and already benefit various applications~\cite{xu2023open,baranchuk2021label,li2023open,tian2023diffuse,zhang2024tale}. 
Such dense annotation prediction also aligns with observations of predicting dense landmarks, helping construct complex non-rigid face patterns~\cite{wood20223d,feng2018joint}. Also, we adopt the coarse-to-fine granularity learning scheme, predicting a series of dense annotations given the sparse head pose coordinates, to better learn the structural geometry of the human face, similar to full body T2I works~\cite{liu2023hyperhuman,ju2023humansd}. 
To support this dense annotation prediction, we collect a large-scale T2I face generation dataset with high-quality images and rich, dense annotations, named T2I-Dense-Face dataset, which also contributes to future face generation research.
In summary, our contributions are:

$\diamond$ We propose a text-to-image diffusion model, named Dense-Face, for personalized image generation. We use the pose-controllable adapter with the pose branch to help the model generate high-fidelity images while maintaining high text controllability. 

$\diamond$ To facilitate Dense-Face to learn domain-specific knowledge for face generation, we devise an auxiliary training objective that predicts a series of dense face annotations. This is supported by our proposed identity-centric text-to-image face dataset, termed T2I-Dense-Face.

$\diamond$ Our proposed method is evaluated on various generation tasks and achieves promising performance in personalization generation and face swapping. 
\Section{Related Works}
\Paragraph{Personalization Diffusion Model}
The personalization T2I diffusion models~\cite{gal2022image_textual_inversion,dreambooth,custom_diffusion,Gligen,controlNet,wei2023elite,gu2023photoswap} attract increasing attentions in these two years. 
FaceX~\cite{han2023generalist} achieves head pose control and identity preservation generation, whereas it has no text-based editing. 
Photomaker~\cite{li2023photomaker} designs a stacked embedding containing a deterministic identity that helps produce exceptional quality customized images. 
InstantID~\cite{wang2024instantid} leverages facial landmarks as the condition, along with the face identity embedding, producing time-efficient and high-fidelity generation. 
These prior works fine-tune the pre-trained SD model on the self-collected face dataset, and such fine-tuning procedures steer the pre-trained SD to the face generation domain, which may decrease the text-based editing ability of the pre-trained SD. 
In contrast, we propose Dense-Face with two generative modes having different functions, which later are combined to generate the final personalized image. In this way, we can learn the face generation domain knowledge while retraining the premium text controllability of the pre-trained SD. 
In addition, we propose a dataset based on the existing face recognition datasets, which provides a sufficient number of images at different poses for each subject, offering more variations in training compared to face samples collected for training in the previous work.

\Paragraph{Learning Specific Generation Domain Knowledge} It is important to equip T2I-SD with domain-specific knowledge for tasks such as face patterns~\cite{gal2023encoder,li2023photomaker,wang2024instantid} and full-body generation~\cite{liu2023hyperhuman,ju2023humansd}, such bio-metric information is important thus widely used~\cite{guo2022multi,guo2019human,abdalmageed2020assessment}.
For this purpose, test-time fine-tuning is first used by works such as Textual inversion~\cite{gal2022image_textual_inversion} and E4T~\cite{gal2023encoder}. Also, the pre-defined condition makes powerful generation guidance like a human skeleton for HyperHuman~\cite{liu2023hyperhuman} and Human SD~\cite{ju2023humansd}. Our work has two differences from prior works. First, we use the more sparse conditions compared to face semantic masks~\cite{huang2023collaborative,cheng2023layoutdiffuse,controlNet,cheng2023consistent,cheng2024rethinking} --- three Euler coordinates (\textit{e.g.}, yaw, pitch and roll), to improve the freedom of the face generation. Secondly, we employ a novel technique of adding domain-specific generation knowledge --- PC-adapter and the pose branch trained to predict dense face annotations from coarse to fine granularities. Note that HyperHuman~\cite{liu2023hyperhuman} adopts such coarse-to-fine granularity learning to train two entire SDs, whereas we only train additional proposed components while keeping SD frozen.
\Section{Method}
\label{sec:method_overall}
Sec.~\ref{sec:problem_formulation} first revisits preliminaries and our problem formulation.
Sec.~\ref{sec:method_architecture} reports the proposed Dense-Face containing the pose branch and the pose-controllable adapter. 
Sec.~\ref{sec:method_dense} and Sec.~\ref{sec:dense_dataset} introduce dense annotation prediction details and our proposed T2I-Dense-Face dataset, respectively. 
Finally, Sec.~\ref{sec:train} and Sec.~\ref{sec:inference} detail training procedures and the latent space blending for the inference, respectively.

\SubSection{Problem Formulation}
\label{sec:problem_formulation}

\Paragraph{Preliminary} 
T2I-SD employs diffusion-and-denoising process in latent space. 
Denote the input target image as 
$\mathbf{I} \in \mathbb{R}^{H\times W\times 3}$. 
First, $\mathbf{I}$ is converted to the latent space feature $\mathbf{Z}_{0}$ by the pre-trained autoencoder $\mathcal{E}$, \textit{i.e.}, $\mathbf{Z}_0 = \mathcal{E}(\mathbf{I})$. 
Given $\mathbf{Z}_{0}$, we can sample $\mathbf{Z}_{t}$ via $\mathbf{Z}_{t} \triangleq \sqrt{\bar{\alpha}_{t}} \mathbf{Z}_{0}+\sqrt{1-\bar{\alpha}_{t}} \bm{\epsilon}$,
where $\bar{\alpha}_{t}{=}\prod_{k=1}^t\alpha_k$ and $\bm{\epsilon} \sim \mathcal{N}(0,\mathbf{I})$, following~\cite{ho2020denoising_ddpm}. The denoising network $\bm{\epsilon_{\theta}}$, commonly structured as a
UNet-based~\cite{ronneberger2015u}, estimates additive Gaussian noise $\bm{\epsilon}$ at the time step $t$:
\begin{equation}
\mathop \mathbb{E} \limits_{t,\mathbf{Z},\bm{\epsilon}, \mathbf{c}}\left[ \left\| \bm{\epsilon} - \bm{\epsilon}_{\theta}\left( \mathbf{Z}_{t}, \mathbf{c}, t \right) \right\| ^2 \right],   \vspace{-1mm}
\label{eq:ldm_loss}
\end{equation} 
where $\mathbf{c}$ is the text caption embedding obtained by 
the pre-trained text encoder with the given text caption.
\begin{table}[t]
    \centering
    \resizebox{1.0\columnwidth}{!}{%
        \begin{tabular}{c|c|l|c|c|l}
            \toprule
            \textbf{Symbol} & \textbf{Dim.} & \textbf{Notation} & \textbf{Symbol}  & \textbf{Dim.} & \textbf{Notation} \\
            \hline \hline
            $\mathbf{I}_{id}$ & Matrix & Reference subject Img.  & $\mathbf{I}_{tar.}$ & Matrix & Reconstructed target Img. \\\hline
            $\mathbf{I}_{pose}$ & Matrix & \shortstack[l]{Img. of head pose representing \\ in $3$ Euler Angles (yaw, roll and pitch).} &$\mathbf{D}$ & Matrix & \shortstack[l]{Img. of the face depth map \\ normalized between [0,1].} \\\hline
            $\mathbf{P}$ & Matrix & \shortstack[l]{Img. of the pseudo mask \\ based on $468$ landmarks}. & $\mathbf{L}$ & Matrix & \shortstack[l]{Img. of the heatmap \\  converted from $468$ landmarks}.\\\hline
            $\mathcal{G}_{1}$ & - & \shortstack[l]{Dense-Face's Text-based mode, \\ identical to pre-trained SD.} & $\mathcal{G}_{2}$ & - & \shortstack[l]{Dense-Face's Face-generation \\mode with PC-adapter, \\$\bm\epsilon_{pose}$, and $\bm\epsilon_{dense}$ enabled.} \\\hline
            $\mathbf{c}^{id}$ & $\mathbb{R}^{d}$ & \shortstack[l]{Deterministic \\ Identity Embedding.} & 
            $\mathbf{c}^{\prime}$ & $\mathbb{R}^{k}$ & \shortstack[l]{Identity \\ Text Embedding.}\\\hline
            $\mathbf{c}$ & $\mathbb{R}^{n\times k}$ & \shortstack[l]{Input Text Embedding \\ $n$ is the token num.} & $\mathbf{f}^{dense}$ & Matrix & \shortstack[l]{SD-UNet feature used \\ predicting the dense annotations.}  \\\hline
            \bottomrule
        \end{tabular}
    }
    \vspace{-3mm}
    \caption{Important symbols and notations. [Key: img: image; num: number.]\vspace{-8mm}}
    \label{tab:symbol}
\end{table}

\Paragraph{Problem Formulation} 
To learn the human face pattern, given the target image $\mathbf{I}_{tar}$, we estimate its head pose images $\mathbf{I}_{pose}$, and dense face annotations including dense landmarks $\mathbf{L}$, the depth map $\mathbf{D}$ and the pseudo face mask $\mathbf{P}$. 
We also select a face image, $\mathbf{I}_{id}$, with the same identity as $\mathbf{I}_{tar}$. 
Therefore, the training sample pair is denoted as $\{\mathbf{I}_{pose}, \mathbf{I}_{id}, \mathbf{c}, \mathbf{I}_{tar}, \mathbf{L}, \mathbf{D}, \mathbf{P}\}$, detailed in Sec.~\ref{subsec:train_sample}. 

As depicted in Fig.~\ref{fig:overview_architecture}, we introduce additional components such as a PC-adapter (${\mathbf{w}^{\prime}}^{q}, {\mathbf{w}^{\prime}}^{v}$, and ${\mathbf{w}^{\prime}}^{v}$), a pose branch ($\bm\epsilon_{pose}$) and an annotation prediction module ($\bm\epsilon_{dense}$) on the top of the pre-trained SD. 
In the \textit{text-editing} mode, the Dense-Face has additional components disabled, the forward propagation of which is then identical to the pre-trained SD. 
We denote such the generation mode as $\mathcal{G}_1$. 
In the \textit{face-generation} mode, the Dense-Face with additional components enabled is represented as $\mathcal{G}_2$. As a result, we have
\begin{equation}
    \mathbf{I} = \mathcal{G}_{1}(\mathbf{c}); 
    \mathbf{L}, \mathbf{D}, \mathbf{P}, \mathbf{I}_{tar.} = \mathcal{G}_{2}(\mathbf{I}_{pose}, \mathbf{I}_{id}, \mathbf{c}). 
    \label{eq:generation_mode}
\end{equation}

Formally, $\mathbf{I}_{tar}$, $\mathbf{I}_{id}$ and $\mathbf{I}_{pose}$ are of the same size, $\mathbb{R}^{H\times W\times 3}$. 
We apply two additional encoders, $\mathcal{E}_{pose}$ and $\mathcal{E}_{id}$, which obtain the pose condition and reference condition, respectively: $\mathbf{c}^{pose} = \mathcal{E}_{pose}(\mathbf{I}_{pose})$ and $\mathbf{c}^{id} = \mathcal{E}_{id}(\mathbf{I}_{id})$. 
Similar to Eq.~\ref{eq:ldm_loss}, we can obtain as the objective function ($\mathcal{L}^{\epsilon}$) for $\bm{\epsilon}_\theta$ in $\mathcal{G}_2$ as:
\begin{equation}
\vspace{-2mm}
\mathcal{L}^{\epsilon}=\mathop \mathbb{E} \limits_{t,\mathbf{Z},\bm{\epsilon}, \mathbf{c}, \mathbf{c}^{pose}, \mathbf{c}^{id}}\left[ \left\| \bm{\epsilon} - \bm{\epsilon}_{\theta}\left( \mathbf{Z}_{t}, \mathbf{c}^{pose}, \mathbf{c}^{id}, \mathbf{c}, t \right) \right\| ^2 \right]. \vspace{-1mm}
\label{eq:face_ldm_loss}
\end{equation}
\subsection{Realistic Face Personalized T2I Diffusion Model}
\label{sec:method_architecture}
This section introduces the pose branch and PC-adapter in our Dense-Face.

\Paragraph{Pose Branch} To achieve pose-controllable face generation, we introduce an additional branch on the top of pre-trained SD, taking as the input the head pose image, which uses Euler angles (\textit{e.g.}, yaw, pitch and roll) representing the head pose.
Such conditions are sparser than the face semantic mask used in prior  works~\cite{huang2023collaborative,cheng2023layoutdiffuse}, providing more freedom in the generation process.
Specifically, this pose branch only leverages the first half of the standard UNet, without the zero-convolution, an effective design proposed in ControlNet~\cite{controlNet}, which reduces harmful impacts from fine-tuning. This is because our pose branch involves major training on the large-scale face dataset (Sec.~\ref{sec:dense_dataset}), and we empirically find the generation performance is similar whether zero-convolution is used or not.

\Paragraph{Text Identity Embedding}
Similar to the previous work~\cite{gu2023photoswap,gal2023encoder-specific-diffusion,wang2024instantid} fusing identity information with the input text embedding, we first convert the identity condition $\mathbf{c}^{id}$ to the \textit{identity text embedding} $\mathbf{c}^{\prime}$. More formally, in the learned faceNet space~\cite{deng2019arcface}, $\mathcal{Q}$, each face identity $\mathbf{c}^{id} \in \mathbb{R}^{d}$ is embedded in the deterministic manifold. 
We use a few Multiple Perception Layer ($\texttt{MLP}$) layers to transform $\mathbf{c}^{id}$ into the text embedding space, $\mathcal{C} \in \mathbb{R}^{k}$, obtained from the pre-trained text encoder. Let us denote this transformed feature as $\mathbf{\Delta}{\mathbf{c}}^{id} \in \mathbb{R}^{k}$,
which can be added onto the pre-trained word embedding ``\texttt{Face}'', a coarse textual descriptor in the face domain generation. Specifically, this process can be described as:
\begin{equation}
    \mathbf{c^{\prime}} = \lambda \mathbf{\Delta}{\mathbf{c}^{id}} + \mathbf{c}^{\texttt{Person}}\\
        = \lambda \cdot \texttt{MLP}(\mathbf{c}^{id}) + \mathbf{c}^{\texttt{Face}},
    \label{eq:id_text_embedding}
\end{equation}
where $\lambda$ controls the magnitude and is empirically set as $1e-2$.
$\mathbf{c^{\prime}}$ is the identity feature in the pre-trained text embedding space, denoted as \textit{identity text embedding}.
This $\mathbf{c^{\prime}}$ is concatenated with the input text condition $\mathbf{c}$ that is input to the Dense-Face. Ideally, given the identity text embedding $\mathbf{c}^{\prime}$,  Dense-Face should generate the image  $\hat{\mathbf{I}}_{tar}$ based on $\mathbf{c}^{id}$, having the identical identity as $\mathbf{I}_{id}$.

\Paragraph{Pose Controllable (PC) Adapter}
With $\mathbf{c}^{\prime}$ as the input, the PC-adapter (${\mathbf{w}^{q\prime}}$, ${\mathbf{w}^{k\prime}}$ and ${\mathbf{w}^{v\prime}}$) modifies the forward propagation of Dense-Face's original cross attention modules, 
helping generation condition on given identity and head pose.
More formally, the cross attention ($\texttt{C-Att}$) module takes general text embedding $\mathbf{c}$ and noisy image latent feature $\mathbf{f}$ as inputs:
\begin{equation}
    \mathbf{f}_{out} = \texttt{C-Att}(\mathbf{q},\mathbf{k},\mathbf{v}) = \texttt{softmax}\left(\frac{\mathbf{q}\mathbf{k}^T}{\sqrt{d_k}}\right) \mathbf{v}, \vspace{-1mm}
    \label{eq:cross_att}
\end{equation}
where $\mathbf{q} = \mathbf{w}^{q} \mathbf{f}$, $\mathbf{k} = \mathbf{w}^{k} \mathbf{c}$ and $\mathbf{v} = \mathbf{w}^{v} \mathbf{c}$ are feature maps for query, key and value, respectively. 
By capturing the correlation between image and text features,  $\texttt{C-Att}$ helps the pre-trained T2I-SD generate images based on text captions.

As depicted in Fig.~\ref{fig:overview_architecture}, the proposed PC adapter alters each $\texttt{C-Att}$ module's forward propagation as:
\begin{equation}
    \mathbf{f^{\prime}}_{out} = \texttt{C-Att}(\mathbf{q}^{\prime},\mathbf{k}^{\prime},\mathbf{v}^{\prime}) = \texttt{softmax}\left(\frac{\mathbf{q}^{\prime}{\mathbf{k}^{\prime}}^T}{\sqrt{d_k}}\right) \mathbf{V}^{\prime}, \vspace{-1mm}
    \label{eq:cross_att_new}
\end{equation}
where ${\mathbf{f}^{\prime}}_{out}$ is the updated output feature of $\texttt{C-Att}$; $\mathbf{q}^{\prime} = (\mathbf{w}^{q} + {\mathbf{w}^{\prime}}^{q}) \mathbf{f}$, $\mathbf{k}^{\prime} = (\mathbf{w}^{k} + {\mathbf{w}^{\prime}}^{k}) \mathbf{c}^{\prime}$ and $\mathbf{v}^{\prime} = (\mathbf{w}^{v} + {\mathbf{w}^{\prime}}^{v}) \mathbf{c}^{\prime}$ are updated feature maps of query, key, and value respectively. $\mathbf{c}^{\prime}$ is the updated embedding comprising of pose, identity and text information. Apparently, PC-adapter generates additional residual features on top of original $\mathbf{q}$, $\mathbf{k}$, $\mathbf{v}$. For example, $\mathbf{q}^{\prime} = \mathbf{w}^{q} \mathbf{f} + {\mathbf{w}^{q\prime}} \mathbf{f} = \mathbf{q} + \Delta \mathbf{q}$. 
Such additional residual features shift the pre-trained T2I-SD to the face generation domain. 

Our proposed pose branch and PC-adapter act as plug-in modules, and Dense-Face can function either with or without them. In other words, Dense-Face has two separate generation modes: \textit{text-editing mode}, where $\texttt{C-ATT}$ outputs $\mathbf{f}_{out}$; \textit{face-generation mode}, where $\texttt{C-ATT}$ outputs $\mathbf{f^{\prime}}_{out}$ that conditions on the input pose $\mathbf{c}^{pose}$ and identity $\mathbf{c}^{id}$. 
These two modes' forward propagation is formulated as Eq.~\ref{eq:generation_mode}.
In inference, we adopt the latent space blending scheme, jointly leveraging both modes, for personalized generation. (see Sec.~\ref{sec:inference}).
\begin{figure}[t]
  \centering
  \includegraphics[width=\linewidth]{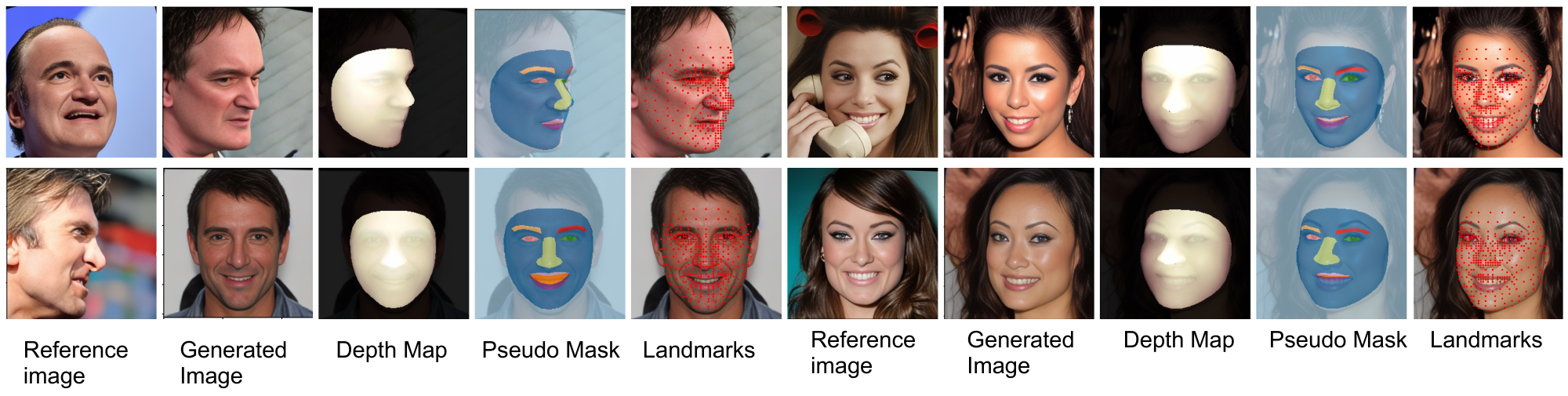}
  \vspace{-5mm}
  \caption{Given the reference image and target head pose (not shown in Figure), the proposed method generates the id-preserved image, depth map, pseudo mask and dense landmark. Such annotation predictions follow the insight that human faces possess inherent structural characteristics across various levels of granularity, encompassing coarse head pose coordinates to fine facial spatial geometry. These annotations are detailed in Sec.~\ref{sec:dense_dataset}. \vspace{-4mm}}
  \label{fig:dense_predict}
\end{figure}
\SubSection{Dense Face Annotation Prediction}
\label{sec:method_dense}

To facilitate Dense-Face's face generation mode (\textit{e.g.}, $\mathcal{G}_2$) in learning face generation domain knowledge, we encourage it to predict dense face annotations, given the head pose image as the condition. Details are shown in Fig.~\ref{fig:dense_predict}.
Unlike the generative process of diffusion models, which requires multiple steps of denoising, the dense prediction only needs a single forward process. 
Formally, given the input pair $\{\mathbf{Z}_{t}, \mathbf{c}, \mathbf{c}_{pose}, \mathbf{c}_{id}\}$, 
we leverage Dense-Face's internal UNet features to predict the $n$ landmarks $\mathbf{L} \in \mathbf{R}^{N \times W \times H}$, the pseudo mask with $m$ semantics $\mathbf{P} \in \mathbf{R}^{m \times W \times H}$ and normalized depth mask $\mathbf{D} \in \mathbf{R}^{1 \times W \times H}$, as depicted in Fig.~\ref{fig:overview_architecture}.
Consequently, we have
\begin{equation}
   {\mathbf{f}_{0}}^{dense} = \bm{\epsilon}_{\theta}(\mathbf{Z}_{t}, \mathbf{c}, \mathbf{c}^{pose}, \mathbf{c}^{id}, t). \vspace{-1mm}
   \label{eq:dense_pred}
\end{equation}

We concatenate these intermediate feature maps output from different UNet blocks, after the necessary upsampling operation. Then, the resultant $3$D tensor are used as the pixel-level representation for the prediction.
This can be described as ${\mathbf{f}_{0}}^{dense}  \longrightarrow
\mathbf{f}^{dense}$. After that, we use the annotation prediction module that contains three individual convolution blocks to predict dense annotations. These predicted landmarks, depth maps, and pseudo masks are denoted as $\mathbf{\hat{L}}$, $\mathbf{\hat{D}}$ and $\mathbf{\hat{P}}$, respectively. 
\SubSection{T2I Dense Face Annotation Dataset (T2I-Dense-Face)}
\label{sec:dense_dataset}
It is important to have a large-scale dataset with high quality samples and rich annotations for T2I face generation methods. Hence, we propose the T2I Dense Face Annotation (T2I-Dense-Face) Dataset, comprising around $2M$ high quality image-text pairs.

\Paragraph{Image Restoration} To construct an identity-centric dataset, we leverage two large-scale face recognition datasets: CASIA-WebFace~\cite{yi2014learning} and CelebA~\cite{liu2018large}. 
We use CodeFormer~\cite{wang2021gfpgan}, 
to restore images from these datasets into a high-quality version, which better fits the need for training a generative model.

\Paragraph{Head Pose}
Given the image, we use Hopenet~\cite{Ruiz_2018_CVPR_Workshops}~\footnote{\href{https://github.com/natanielruiz/deep-head-pose}{https://github.com/natanielruiz/deep-head-pose}} to estimate yaw, row and pitch scores, and convert these scores into a 3D coordinate image.  

\Paragraph{Landmark and Pseudo Face Mask}
According to the previous work~\cite{wood20223d}, dense face alignment helps learn face patterns, 
so we use Google mediapipe~\footnote{\href{https://github.com/google/mediapipe}{https://github.com/google/mediapipe}} to dump $468$ landmarks for our dataset. Moreover, we notice that some landmarks are more important to capture the face pattern, so we create the pseudo face mask based on estimated dense landmarks to outline different facial areas (Fig.~\ref{fig:dense_predict}), such as the face shape, nose shape and eyebrow length.

\newcommand{\jiaxin}[1]{\textcolor{cyan}{#1}}
\Paragraph{Depth Map} We also utilize depth information predicted by Google mediapipe to generate a 3D facial depth map, providing enhanced geometric supervision during the training of Dense-Face.

\begin{figure}[t]
  \centering
  \includegraphics[width=1\linewidth]{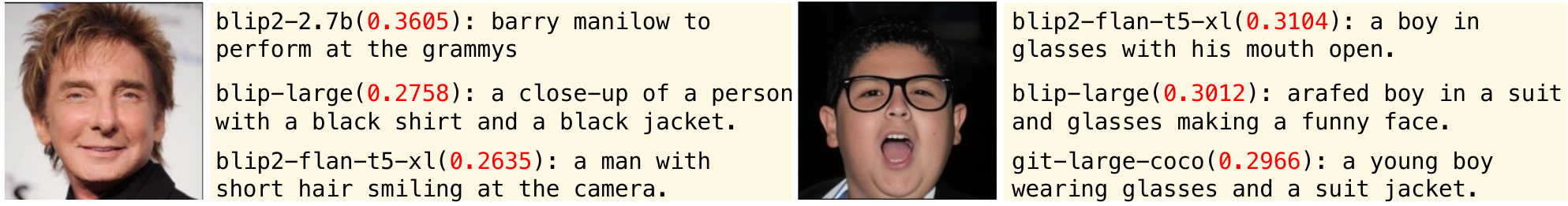}
  \vspace{-4mm}
  \caption{
  We use various image caption methods via the third-party implementation~\cite{wolf2019huggingface} to generate captions. These captions have different text-image alignment scores. We only include captions with three highest \textcolor{red}{scores} in training samples. \vspace{-3mm}}
  \label{fig:dataset}
\end{figure}

\Paragraph{Image Caption} 
Given an image, we use different caption generation methods~\cite{li2022blip,li2023blip,zhang2022opt,chung2022scaling} to generate captions (Fig.~\ref{fig:dataset}). To ensure image-to-text variations and accuracy, we only keep captions with three highest scores in the training. 
\label{subsec:train_sample}

\SubSection{Training}
\label{sec:train}
\Paragraph{Training}
In the training, in addition to  $\mathcal{L}^{\epsilon}$ (Eq.~\ref{eq:face_ldm_loss}), we jointly minimize three losses that predict the dense annotation.
First, following the face alignment work~\cite{bulat2017far,tang2018quantized,tang2018quantized}, we convert dense landmarks into a heatmap image $\hat{\mathbf{L}}$ as the ground truth. Then we minimize $\emph{l}_{2}$ distance between the $i^{th}$ predicted heatmap ($\hat{\mathbf{L}_{i}}$) and the corresponding ground truth ($\mathbf{L}_{i}$), 
denoted as $\mathcal{L}^{LD}$. 
For the pseudo face mask,  $\mathcal{L}^{PM}$ computes the cross entropy (\texttt{CE}) between $i^{th}$ pseudo masks $\mathbf{P}_{i}$ and $\hat{\mathbf{P}}_i$. 
For the face depth mask, we first normalize the value to $[0, 1]$ and then compute the $\emph{l}_{2}$ distance between $\mathbf{D}$ and $\hat{\mathbf{D}}$. Then, we have the overall loss where $\lambda_{1}$, $\lambda_{2}$, $\lambda_{3}$ are hyper-parameters, 
\begin{equation}
    \mathcal{L}^{total} = \mathcal{L}^{\epsilon} + 
    \underbrace{\lambda_{1} \frac{1}{M} \sum_{i=1}^{M} \left\| \mathbf{L}_{i} - \hat{\mathbf{L}_{i}} \right\| ^2}_{\mathcal{L}^{LD}} +
    \underbrace{\lambda_{2} \frac{1}{N} \sum_{i=1}^{N}\texttt{CE} (\mathbf{P}_i, \hat{\mathbf{P}}_{i})}_{\mathcal{L}^{PM}} +
    \underbrace{\lambda_{3} \frac{1}{N} \sum_{i=1}^{N}\left\| \mathbf{D}_i -  \hat{\mathbf{D}}_{i}\right\| ^2}_{\mathcal{L}^{DE}}.
    \label{eq:total}
\end{equation}
\begin{figure}[t]
    \begin{minipage}{0.48\textwidth}
    \begin{subfigure}{1.\textwidth}
    \includegraphics[width=1\linewidth]{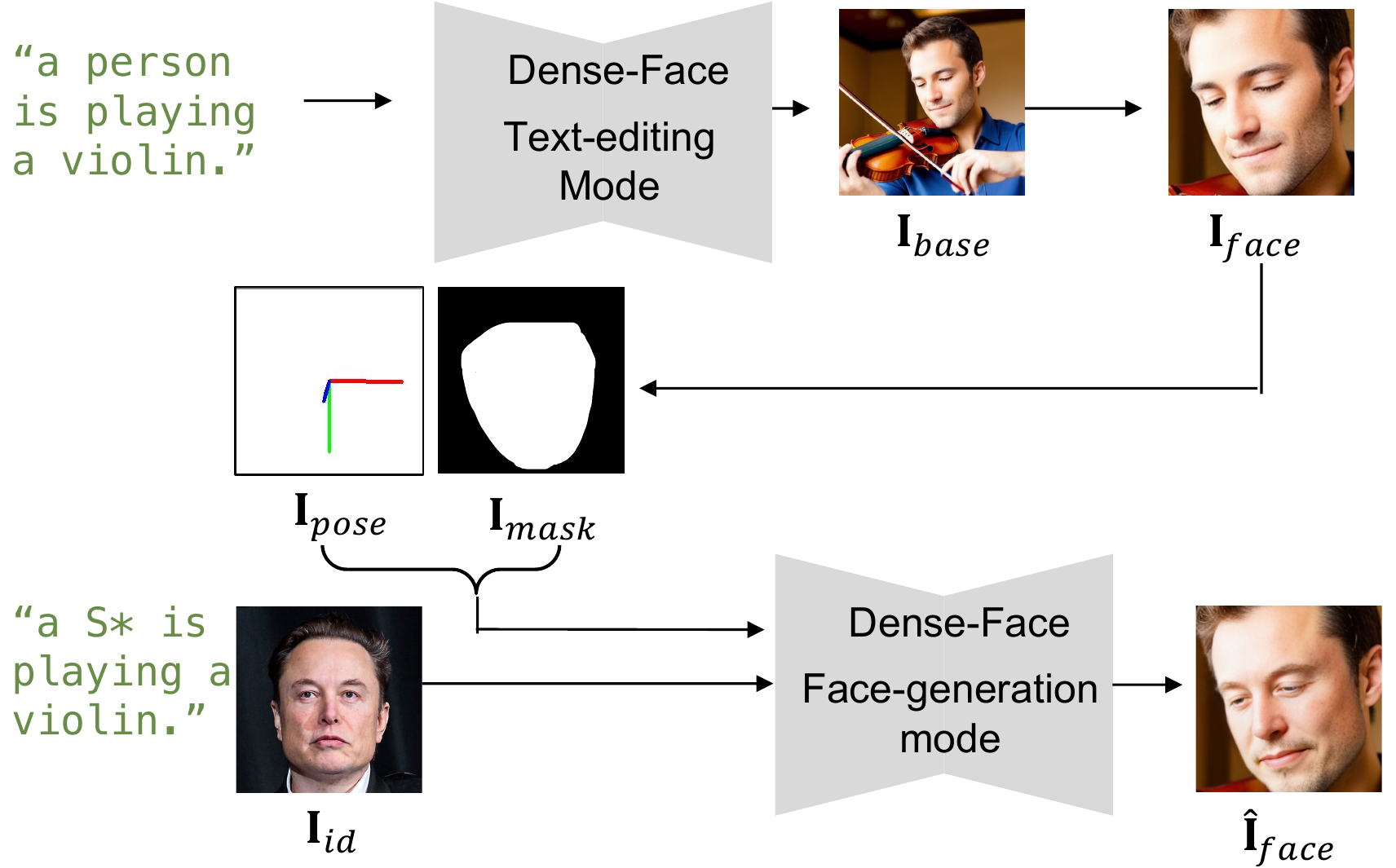}
    \caption{Dense-Face's text-editing mode ($\mathcal{G}_{1}$) generates the base image (\textit{i.e.}, $\mathbf{I}_{base}$), which contains diverse contexts, and its face region is cropped as $\mathbf{I}_{face}$. Then the face generation mode ($\mathcal{G}_{2}$) uses the latent blending algorithm (Fig.~\ref{fig:blending_block}) to modify the $\mathbf{I}_{face}$ into the $\widehat{\mathbf{I}}_{face}$ with a consistent identity as the given subject of $\mathbf{I}_{id}$.}
    \label{fig:blending_procedure}
    \end{subfigure}
    \begin{subfigure}{1.\textwidth}
    \includegraphics[width=1\linewidth]{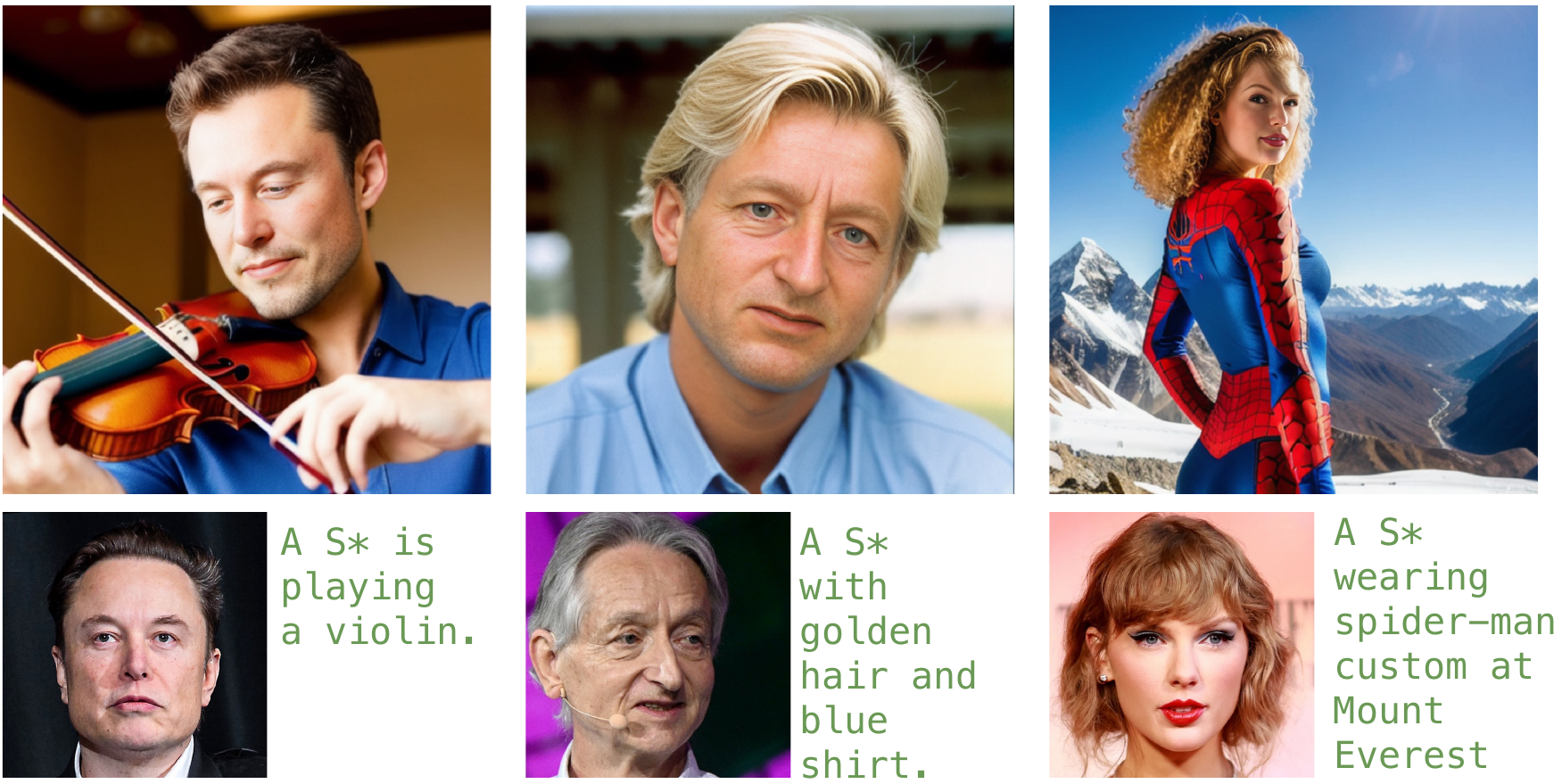}
    \caption{The latent blending image result of people in different contexts and at different pose angles. \vspace{-3mm}}
    \label{fig:blending_result}
    \end{subfigure}
    \end{minipage}
    \begin{minipage}{0.48\linewidth}
        \begin{algorithm}[H]
        \begin{algorithmic}
            \scriptsize
            \caption*{\small Latent Space Blending}
            \STATE \textbf{Input:} cropped image $\mathbf{I}_{face}$, reference  image $\mathbf{I}_{id}$, face  mask $\mathbf{I}_{mask}$, head pose image $\mathbf{I}_{pose}$, text embedding $\mathbf{c}$ and diffusion steps $k$.
            \STATE \textbf{Output:} updated face image $\widehat{\mathbf{I}}_{face}$ that has the consistent identity with $\mathbf{I}_{id}$.
            \STATE \textbf{Models:} Face-Dense in the face generation mode $\mathcal{G}_{2}$, containing VAE = ($\mathcal{E}(\mathbf{I})$, $\mathcal{D}(\mathbf{z})$) and Diffusion Model $\bm{\epsilon}_{\theta}$ = (\textit{noise}($\mathbf{c}, t$), \textit{denoise}$(\mathbf{z}^{t}, \mathbf{c},\mathbf{c}^{id}, \mathbf{c}^{pose}, t)$) \COMMENT{defined in Eq.~\ref{eq:face_ldm_loss}}.
            
            \STATE $\mathbf{z}^{init.}_{bg} = \mathcal{E}(\mathbf{I}_{face})$
            \texttt{//\textcolor{blue}{Convert to the latent space as the background image.}}
            \STATE $\mathbf{c}^{id} = \mathcal{E}_{id}(\mathbf{I}_{id})$; 
            $\mathbf{c}^{pose} = \mathcal{E}_{pose}(\mathbf{I}_{pose})$ 
            
            \texttt{//\textcolor{blue}{Initialize conditions.}} 
            \STATE $\mathbf{I}^{latent}_{mask}$ = \textit{downsample}$(\mathbf{I}_{mask})$
            \FOR{$t$ from $k$ to $0$}   
                \IF{$t$ == $k$}
                    \STATE $\mathbf{z}^{t} \gets \textit{noise}(\mathbf{z}^{init.}_{bg}, k)$
                \ENDIF
                \STATE $\mathbf{z}^{t}_{fg} \gets \textit{denoise}(\mathbf{z}^{t}, \mathbf{c},\mathbf{c}^{id}, \mathbf{c}^{pose}, t)$  
                
                \texttt{//\textcolor{blue}{Face image latent features.}}
                \STATE $\mathbf{z}^{t}_{bg} \gets \textit{noise}(\mathbf{z}^{init.}_{bg}, t)$
                
                \texttt{//\textcolor{blue}{Background image latent features.}}
                \STATE $\mathbf{z}^{t} \gets \mathbf{z}^{t}_{fg} \odot \mathbf{I}^{latent}_{mask} + \mathbf{z}^{t}_{bg} \odot (1 - \mathbf{I}^{latent}_{mask})$ \texttt{//\textcolor{blue}{Latent blending based on the mask.}} 
            \ENDFOR
            \STATE $\widehat{\mathbf{I}}_{face} \gets \mathcal{D}{(\mathbf{z}^{0})}$
            \RETURN $\widehat{\mathbf{I}}_{face}$
        \end{algorithmic}
        \end{algorithm}
        \vspace{-10mm}
        \caption{Algorithm Block}
        \label{fig:blending_block}
    \end{minipage}
    \vspace{-3mm}
    \caption{The latent space blending.\vspace{-3mm}}
\end{figure}
\vspace{-2mm}
\SubSection{Inference}
\label{sec:inference}
We adapt our proposed method for the personalization T2I generation via the common latent space blending~\cite{avrahami2023blended,liu2024towards,zhang2023paste,avrahami2022blended}. 
First, we apply \textit{text-editing mode}, $\mathcal{G}_1$, to generate the base image $\mathbf{I}_{base}$ that contains a subject in rich contexts (as depicted in the Fig.~\ref{fig:blending_procedure}). 
Then, we crop the face region of $\mathbf{I}_{base}$ as $\mathbf{I}_{face}$, on which we obtain the head pose $\mathbf{I}_{pose}$ and the binary mask $\mathbf{I}_{mask}$ of the face region. 
Secondly, we apply the face generation mode, $\mathcal{G}_2$, taking the text-editing mode's output along with the reference image $\mathbf{I}_{id}$ and text caption $\mathbf{c}$ for generating the new face image $\widehat{\mathbf{I}}_{face}$ with the same identity as $\mathbf{I}_{id}$. Specifically, $\mathbf{I}_{mask}$ helps blend the latent space feature of $\mathbf{I}_{face}$ with $\mathcal{G}_2$'s diffusion process, which re-paints the face region of $\mathbf{I}_{face}$ while preserving its background. Lastly, we paste $\widehat{\mathbf{I}}_{face}$ back to $\mathbf{I}_{base}$ for final personalized results. 
The mathematical equation is in the Algorithm block of Fig.~\ref{fig:blending_block}. 
As shown in Fig.~\ref{fig:blending_result}, the blending process completes in the latent space, which largely reduces artifacts in the RGB domain. Because of this property, latent space blending can result in high-fidelity images and has been widely adopted~\cite{avrahami2023blended,liu2024towards,zhang2023paste,avrahami2022blended}.
\Section{Experiment}
\label{sec:exp}
\Paragraph{Experiment Setup} We train Dense-Face on T2I-Dense-Face that contains $2M$ image-text pairs with around $20k$ identities. For a fair comparison, we use $30$ unseen celebrities to evaluate different methods. 
Personalized image generation is a challenging task that requires evaluation on diverse aspects, so we choose the following metrics: (1) \textit{Image Fidelity}, we use CLIP-I~\cite{gal2022image_textual_inversion} and DINO~\cite{caron2021emerging} scores to measure the fidelity between generated and real images. (2) \textit{Text Alignment}: CLIP-T~\cite{radford2021learning} measures the model's text controllability. Both CLIP-T and CLIP-I use the pre-trained CLIP feature space to evaluate the alignment between images as well as images to their given text captions, as in the prior work~\cite{dreambooth,custom_diffusion,gu2023photoswap}. (3) \textit{Identity Consistency}: 
We use identity similarity as the indicator, which computes the cosine similarity between identity embeddings of generated and real images. In case the proposed method becomes overfitting on the identity embedding from the Arcface, we employ the face embedding obtained from both Arcface~\cite{deng2019arcface} or Adaface~\cite{kim2022adaface} for the measurement. (4) \textit{Face Diversity}: we follow the previous work~\cite{gu2023photoswap}, using  LPIPS scores to measure the diversity of generated facial regions, (5) \textit{Image Quality}: we compute the Fréchet Inception Distance (FID) score between the generated image from different methods with $10,000$ from the FFHQ~\cite{karras2019style}. The motivation for using FFHQ instead of LAION5B is we notice that FFHQ contains better image quality than the majority of images from LAION5B that are collected from web pages. 

\Paragraph{Baselines} We first compare with two general T2I methods: \textit{Stable Diffusion}~\cite{rombach2022high} (SD) and \textit{Control-Net}~\cite{controlNet}. For the SD, we examine three versions of SD (\textit{e.g.}, $1.5$, $2.1$ and XL), and input captions contain the celebrity name for the generation. 
We use the head pose image as the generation condition for the ControlNet, and train such a ControlNet on our T2I-Dense-Face.
Additional baselines include the SoTA personalized diffusion methods, e.g., \textit{IP-adapter} (IPA) and \textit{IP-adapter-Face-ID-plus}~\cite{ye2023ip-adapter} (IPA-FaceID++) which take the image and learned identity feature as the input; \textit{PhotoMaker}~\cite{li2023photomaker} and \textit{InstantID}~\cite{wang2024instantid}, two recent methods that achieve exceptional customized generation performance.

\Paragraph{Implementation Details} We build up our Dense-Face based on pre-trained Stable Diffusion $2.1$, and the pose branch is based on the publicly available ControlNet. Specifically, we train the Dense-Face on T2I-Dense-Face with a batch size of $12$ and a learning rate of $5e-6$ on $8$ NVIDIA RTX A$6000$ GPUs for $3.5$ weeks. The $25$ steps DDIMScheduler with an improved noise schedule is used, and all images are cropped and resized into $512 \times 512$.
\SubSection{Main Generation Performance}
\label{sec:main_peformance}
Tab.~\ref{tab:main_result} reports different methods' personalized performances. 
First, Dense-Face achieves the best FID score, indicating that Dense-Face can generate high-fidelity images. We believe this advantage is from effective face generation domain knowledge learned via the PC adapter trained via dense annotation predictions.
Also, Dense-Face achieves the best text-alignment score (CLIP-T) that is better than the previous personalized models ($0.10$ and $0.13$ higher than photomaker and InstantID, respectively) and comparable with the SDXL, which is the more advanced version of stable diffusion. This indicates that Dense-Face leverages two generation modes to help maintain the premium text controllability.

More formally, in the first three rows, three versions of SD are general text-to-image generation models, achieving high text-alignment scores. However, these pre-trained SDs cannot generate a satisfactory identity-preserved image, even after adding the celebrity name in the text caption, indicating the importance of personalized T2I diffusion models for the subject's customized generation. 
After that, ControlNet in line $\#4$ generates images consistent with the given target head pose.
However, it fails to deliver identity-consistent generation, achieving identity consistency scores of $0.012$ and $0.032$ when measured via embeddings from the pre-trained Arcface and Adaface.

In lines, $\#5$ and $\#6$, IPA and IPA-FaceID++ are two powerful methods that take multiple face conditions to guide the generation. IPA-FaceID++ achieves the best DINO score of $0.643$ and competitive identity preservation performance (its face similarities score is lower than ours and InstantID). 
In lines, $\#7$ and $\#8$, Photomaker and InstantID achieve competitive personalized generation performance. For example, InstantID has the highest identity preservation score of $0.574$ and $0.611$. 
However, InstantID still cannot make an ideal personalized method since it achieves the lowest CLIP-T score, which means its generation cannot effectively condition the input text. 
This aligns with our observation that InstantID does not have accurate generation on objects like "hats", and the generated image always has similar view angles as the input reference image. This is supported by the fact its face diversity score ($0.556$) is lower than other methods. 
Photomaker generates images with competitive performance on both image and text alignment scores, with comparable CLIP-I than InstantID but $0.035$ higher on CLIP-T score. 
Lastly, except the exceptional FID and CLIP-T scores, Dense-Face achieves the best CLIP-I score and second-best face similarities. This superiority showcases that Dense-Face learns the face generation domain knowledge and enhances the generation performance on both identity preservation and image quality.
\begin{table}[t]
\centering
\resizebox{1\textwidth}{!}{
        \begin{tabular}{l|ccccccccc}
            \toprule
            &\textbf{Model} & FID $\downarrow$ & CLIP-T  $\uparrow$ &  CLIP-I  $\uparrow$  & DINO $\uparrow$ &  \shortstack[c]{Arcface\\ID Sim.}$\uparrow$& \shortstack[c]{Adaface\\ID Sim.}$\uparrow$& \shortstack[c]{Face\\Div.}$\uparrow$\\ \midrule
            $1$&SD $1.5$  &$103.93$& $0.330$ &$0.688$&  $0.535$& $0.348$&$0.313$&$\mathbf{\textcolor{red}{0.795}}$\\ 
            $2$&SD $2.1$ &$102.03$& $0.339$ & $0.705$ & $0.535$ &$0.354$&$0.306$&$\mathbf{\textcolor{blue}{0.747}}$\\ 
            $3$&SD-XL  &$116.55$& $\mathbf{\textcolor{red}{0.343}}$ & $0.695$ & $0.577$ & $0.354$&$0.306$&$0.670$\\ 
            $4$&ControlNet &$113.33$& $0.330$ & $0.712$ &$0.489$&$0.012$&$0.032$&$0.726$\\
            \hline
            $5$&IPA &$108.93$& $0.258$ & $0.698$ & $0.554$ &$0.383$&$0.337$&$0.676$\\
            $6$&IPA-FaceID++&$127.20$& $0.263$ & $0.711$ &$\mathbf{\textcolor{red}{0.643}}$&$0.501$&$0.421$&$0.689$\\
            $7$&InstantID &$\mathbf{\textcolor{blue}{97.53}}$ & $0.210$ & $\mathbf{\textcolor{blue}{0.721}}$ & $\mathbf{\textcolor{blue}{0.629}}$ &$\mathbf{\textcolor{red}{0.611}}$&$\mathbf{\textcolor{red}{0.574}}$&$0.556$\\
            $8$&PhotoMaker &$107.39$&$0.245$ & $0.720$ & $0.562$  &$0.366$&$0.324$&$0.694$\\
            \hline
            $9$&Dense-Face (ours) &$\mathbf{\textcolor{red}{95.70}}$& $\mathbf{\textcolor{red}{0.343}}$ & $\mathbf{\textcolor{red}{0.727}}$ & $0.569$ &$\mathbf{\textcolor{blue}{0.568}}$&$\mathbf{\textcolor{blue}{0.519}}$&$0.690$\\ \bottomrule
        \end{tabular}
    }
\caption{Personalized generation performance. [Key: 
\textbf{\textcolor{red}{Bold}}: best; \textbf{\textcolor{blue}{Bold}}: second best]\vspace{-5mm}}
\label{tab:main_result}
\end{table}
\begin{figure}[t]
    \centering
    \includegraphics[width=1.\linewidth]{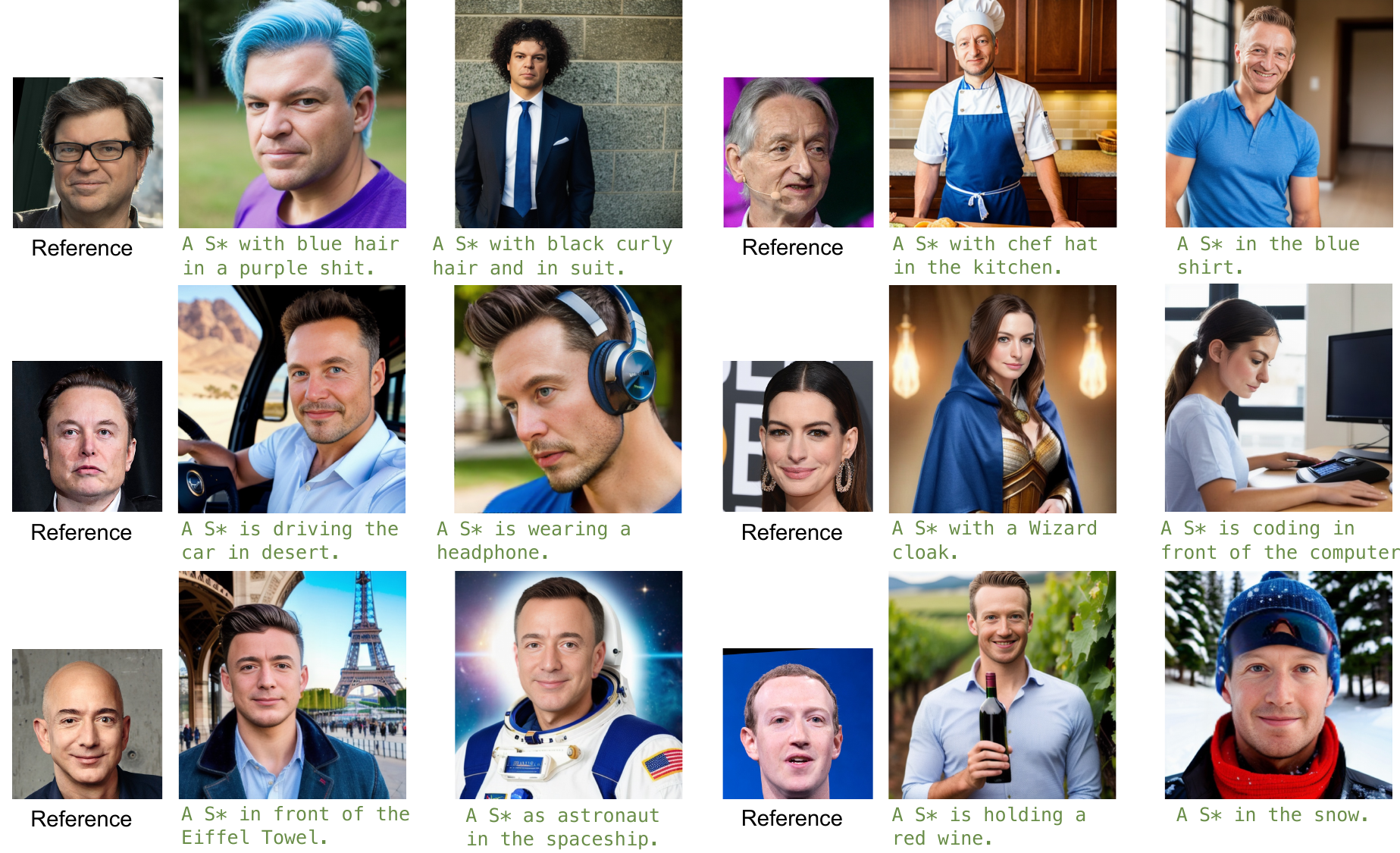}
    \vspace{-7mm}
    \caption{Dense-Face can place subjects in diverse contexts with changed attributes, such as hair color and clothes. \vspace{-6mm}}
    \label{fig:face_showcase_v2}
\end{figure}
\SubSection{Ablation Study and Analysis}
Tab.~\ref{tab:ablation_study} ablates Dense-Face on the test set of $30$ celebrity subjects, and each model is trained with $6,000$ subjects from T2I-Dense-Face. 
Aside from the ID consistency score and FID score that indicates the generation quality, we also report the pose accuracy in Mean Absolution Error (MAE) between the condition head pose and the one estimated from a generated image via HopeNet~\cite{Ruiz_2018_CVPR_Workshops}.

\Paragraph{Dense-Face Architecture} First, compared to the full model, removing the PC-adapter leads to an increase in the FID score by $8.90$ and a decrease in the identity consistency score by $0.207$, as reported in the second row.
This indicates the PC adapter helps identity preservation and the overall generation quality. 
Moreover, when we remove both the PC-adapter and ID branch, the identity similarity score further reduces by around $0.06$ compared to the second row. 
However, we obtain the best pose estimation accuracy of $5.71$ in this setting. We hypothesize this is because the proposed method becomes more sensitive to the pose condition when it only has the pose branch. 
In addition, without the pose branch, MAE increases by $17.20$, which is vastly worse than the full model in pose estimation. 
In the last row, we evaluate the contribution of three regularization terms used in Eq.~\ref{eq:total}. It is observed that the FID score increases by $7.0$ compared to the full model. This result supports our claim that the dense loss assists the proposed method in gaining knowledge of the face generation domain.

\begin{table}[t]
    \begin{subtable}[b]{0.6\linewidth}
        \footnotesize
        \centering
        \resizebox{0.95\textwidth}{!}{
            \begin{tabular}{l|ccc}
             \toprule
               \textbf{Model} & \textbf{FID} $\downarrow$ &  \shortstack[c]{ID\\Similarity}  $\uparrow$ & \shortstack[c]{Pose\\Accuracy} $\downarrow$ \\ \midrule
             Full  & $\mathbf{100.3}$ & $\mathbf{0.357}$ & $5.83$  \\ \hline
             w/o PC-ada & $109.2$ & $0.151$ & $5.74$   \\ 
             w/o (PC-ada + ID bran.) & $113.3$ & $0.093$ & $\mathbf{5.71}$  \\
             w/o pose bran. & $115.4$ & $0.321$ & $22.91$ \\
             w/o dense regularization & $107.3$ & $0.349$  & $5.85$ \\  \bottomrule
            \end{tabular}
        }
        \caption{}
        \label{tab:ablation_study}
    \end{subtable}
    \begin{subfigure}[b]{0.38\linewidth}
    \centering
        \resizebox{0.97\textwidth}{!}{
            \includegraphics[width=\linewidth]{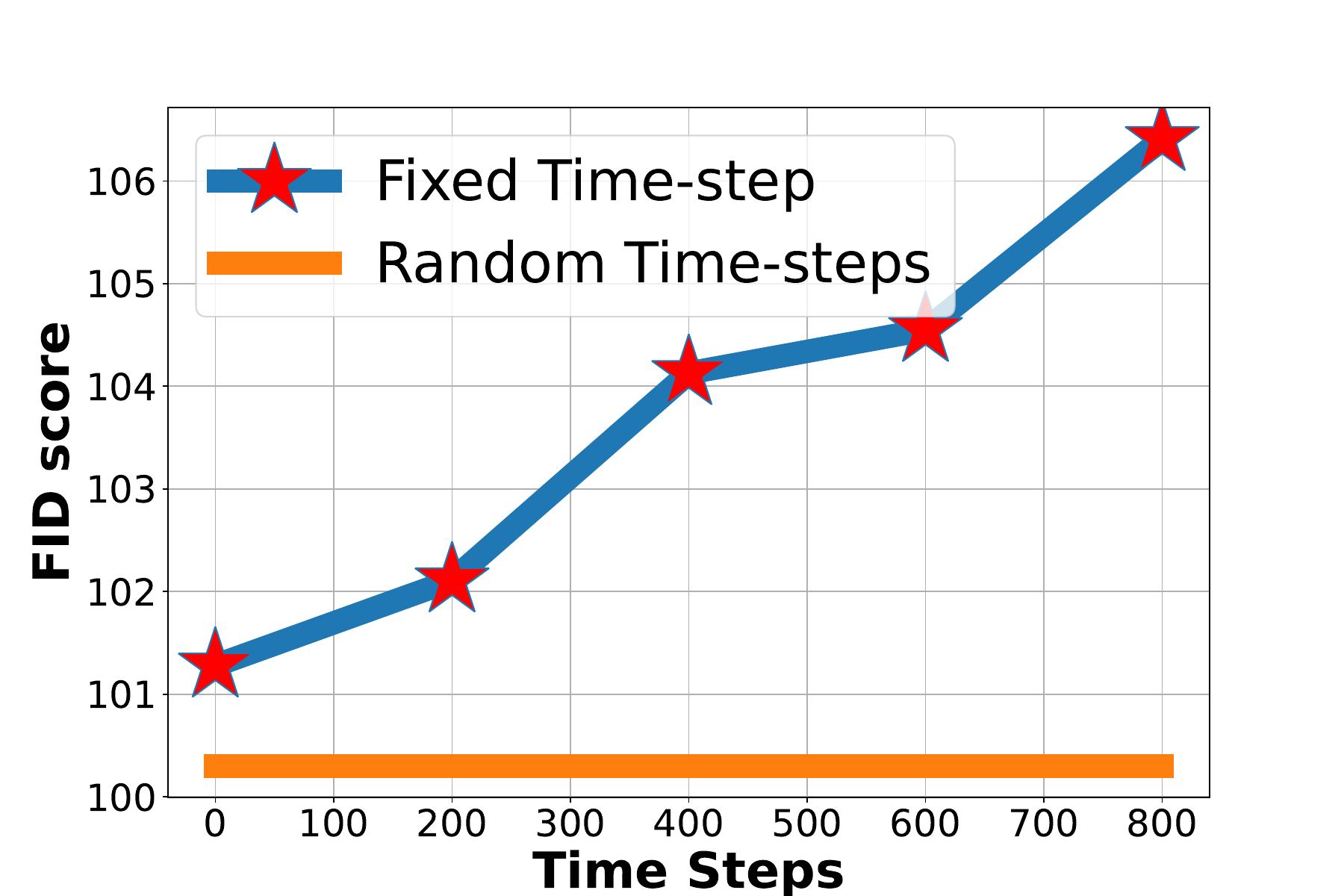}
        }
        \caption{}
        \label{fig:time_step}
    \end{subfigure}
    \vspace{-2mm}
    \caption{(a) Ablation study [\textbf{Bold}: Best.]. (b) Analysis of using different diffusion time steps for the dense annotation prediction. \vspace{-6mm}}
\end{table}

\Paragraph{Diffusion Time-step} 
In Fig.~\ref{fig:time_step}, we use different diffusion time steps (\textit{i.e.}, $t$) for dense annotation prediction (Eq.~\ref{eq:dense_pred}) to determine the most effective time step for the generation. 
Specifically, with a total of 1,000 diffusion steps, we evenly select different steps (\textit{e.g.}, 0, 200, 400, 600, 800) for comparison. As depicted in Fig.~\ref{fig:time_step},
the FID score increases from smaller $t$ to larger $t$, which aligns with the observation that larger $t$ values introduce greater noise distortions to the original image latent space. This is in accordance with the noise process defined in DDPM~\cite{ho2020denoising_ddpm}. 
However, none of these time steps can provide more effective assistance than randomly sampling $t$ for the final generation performance.

\Paragraph{Face Swap Application}
\begin{table}[t]
    \begin{subtable}[b]{0.44\linewidth}
        \footnotesize
        \centering
        \resizebox{1.\textwidth}{!}{
            \begin{tabular}{l|ccc}
             \toprule
               \textbf{Method} & \begin{tabular}[c]{@{}c@{}}ID\\Retrieval\end{tabular}$\uparrow$ & \begin{tabular}[c]{@{}c@{}}Pose\\Accuracy\end{tabular}$\downarrow$ \\ \midrule
                FaceSwap~\cite{rossler2019faceforensics++v3} & $54.19\%$ & $2.51$   \\ 
                Faceshifter~\cite{li2019faceshifter} & $97.38\%$ & $2.96$   \\
                Simswap~\cite{chen2020simswap} & $92.83\%$ & $1.53$  \\
                HiFi-Face~\cite{wang2021hififace} & $98.48\%$ & $5.06$ \\ 
                \hline
                Face-dense & $95.47\%$ &$1.87$ \\  \bottomrule
            \end{tabular}
        }
        \caption{}
        \label{tab:face_swap}
    \end{subtable}
    \begin{subfigure}[b]{0.54\linewidth}
    \centering
        \begin{overpic}[width=1\linewidth]{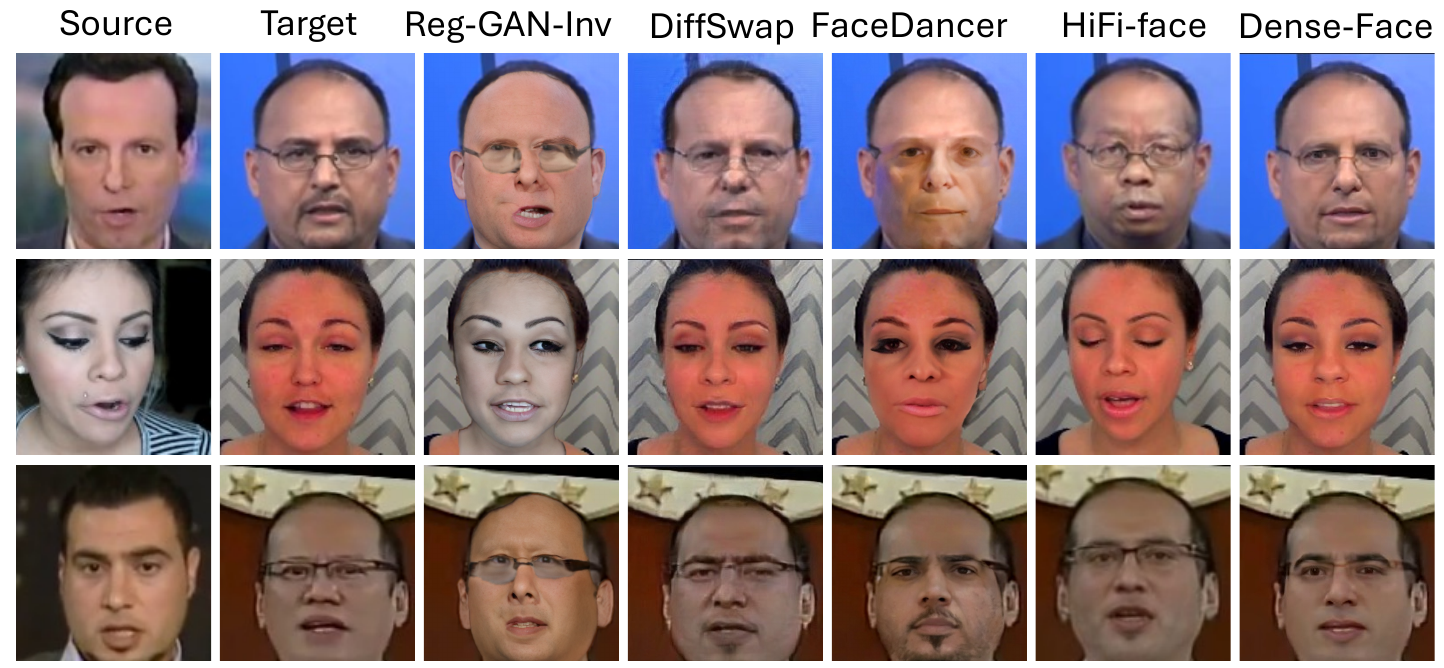}
        \put(55,60.5){\tiny{\cite{li2019faceshifter}}}
        \put(74.5,60.5){\tiny{\cite{wang2021hififace}}}
        \end{overpic}
        \caption{
        }
        \label{fig:face_swap}
    \end{subfigure}
    \caption{(a) Face swap result. We follow the previous work on measuring the face swap algorithm in identity retrieval rate and pose accuracy. (b) The identity feature is learned from the source image and used to modify the face region of the target image. \vspace{-10mm}}
    \label{fig:face_major}
\end{table}
We conduct the evaluation on the face swapping task to demonstrate that the proposed method has learned mapping from the deterministic face identity embedding to faces in the RGB domain. 
Specifically, we feed Dense-Face with identity embedding of the source face, ``re-painting'' the face region of the target image. The quantitative and qualitative results are shown in Tab.~\ref{fig:face_major}, which demonstrate our method achieves competitive face swapping results as prior face swapping methods. After all, in the previous work,
all model parameters are solely trained for face swapping purposes. 
In contrast, Dense-Face only optimizes the additional components and maintains the original text-controllability of the pre-trained SD.
\Section{Conclusion}
We propose a text-to-image diffusion model called Dense-Face for the personalized generation. Being orthogonal to the previous work, our proposed method leverages the pose branch and PC-adapter to make the algorithm have two different generation modes, which are jointly used for the final personalized image. This algorithm design helps achieve high fidelity and exceptional text controllability.
Moreover, we have collected the first text-to-image human face dataset with dense face annotations, which can benefit future research in face generation. 

\Paragraph{Broader Impact} 
We advocate for the community to work towards mitigating potential negative societal impacts.
For example, generated face images could be misused to propagate biases or spread misinformation.
To address this, users can detect these attacks using existing face deepfake detection methods~\cite{xiao_hifi_net++,deepfakeVQA,xiao_neurips,hifi_net_xiaoguo,RED_journal,xiufeng_lamma_detection,pan2024towards} or by training one using images generated by Dense-Face.
\clearpage
\clearpage
\setcounter{page}{1}
\title{Dense-Face: Personalized Face Generation Model via Dense Annotation Prediction \\ -- Supplementary --} 
\author{Xiao Guo\inst{1}\orcidlink{0000-0003-3575-3953} \and
Manh Tran\inst{1} \and
Jiaxin Cheng\inst{2} \and
Xiaoming Liu\inst{1} \orcidlink{0000-0003-3215-8753}
}

\authorrunning{X. Guo et al.}

\institute{Michigan State University \and
University of Southern California, Information Sciences Institute
\email{\{guoxia11,tranman1,liuxm\}@msu.edu}, \email{chengjia@isi.edu}\\
\href{https://chelsea234.github.io/Dense-Face.github.io/}{Dense-Face.github.io}
}
\maketitle
\setcounter{section}{0}
\setcounter{table}{0}
\setcounter{figure}{0}

\noindent In this supplementary, we provide:

$\diamond$ The detailed experiment setup.

$\diamond$ Additional personalized generation results and comparisons.

$\diamond$ Additional visualizations on the T2I-Dense-Face dataset and dense annotation predictions.

$\diamond$ Additional stylization results of Dense-Face.

$\diamond$ Additional face swapping visualizations on the FF++ dataset.

\section{Experiment Setup}
\Paragraph{Evaluation Data} Table.~\ref{tab:evaluate_ids} of the supplementary reports celebrity names that we used for the evaluation in the Sec.~\ref{sec:exp} in the main paper. Please note all these celebrities are unseen subjects during training --- we deliberately remove subjects like "Bruce Li" from the training set to avoid data leakage. For a fair comparison, we use $40$ different text captions from the previous work~\cite{li2023photomaker} to compute the result reported in Tab.~\ref{tab:main_result} of the main paper. Specifically, for each subject, we use multiple images ($4$ to $15$) collected from the website. Each image is paired with all text captions for the personalized generation. 

\Paragraph{Metrics} Empirically, we apply these open-source codes to compute various metrics used in the experiment section. 

$\diamond$ FID: \href{https://github.com/mseitzer/pytorch-fid}{https://github.com/mseitzer/pytorch-fid}

$\diamond$ CLIP-I: \href{https://github.com/rinongal/textual_inversion/tree/main/evaluation}{https://github.com/rinongal/textual\_inversion/evaluation}

$\diamond$ CLIP-T:
\href{https://github.com/rinongal/textual_inversion/tree/main/evaluation}{https://github.com/rinongal/textual\_inversion/evaluation}

$\diamond$ DINO: \href{https://github.com/facebookresearch/dino}{https://github.com/facebookresearch/dino}

$\diamond$ Face similarity (arcface): \href{https://github.com/deepinsight/insightface/tree/master/recognition/arcface_torch}{https://github.com/deepinsight/insightface/tree\\/master/recognition/arcface\_torch}

$\diamond$ Face similarity (adaface): \href{https://github.com/mk-minchul/AdaFace}{https://github.com/mk-minchul/AdaFace}

$\diamond$ Face diversity: \href{https://github.com/richzhang/PerceptualSimilarity}{https://github.com/richzhang/PerceptualSimilarity}

\Paragraph{Baselines} For various personalized methods, we leverage following open-source implementations for comparison.

$\diamond$ Stable Diffusion ($1.5$, $2.1$ and $XL$): \href{https://huggingface.co/docs/diffusers/index}{https://huggingface.co/docs/diffusers}

$\diamond$ ControlNet: \href{https://github.com/lllyasviel/ControlNet}{https://github.com/lllyasviel/ControlNet}

$\diamond$ IP-adapter:
\href{https://github.com/tencent-ailab/IP-Adapter}{https://github.com/tencent-ailab/IP-Adapter}

$\diamond$ IP-adapter-Face++:
\href{https://github.com/tencent-ailab/IP-Adapter}{https://github.com/tencent-ailab/IP-Adapter}

$\diamond$ Photomaker
\href{https://photo-maker.github.io/}{https://photo-maker.github.io/}    

$\diamond$ InstantID
\href{https://github.com/InstantID/InstantID}{https://github.com/InstantID/InstantID}

\begin{table}[t]
    \centering
    \begin{tabular}{lll}
    \toprule
        \multicolumn{3}{c}{Evaluation IDs}\\
    \midrule
   Alan Turing &Fei-Fei Li&Michelle Obama\\
    Albert Einstein &Geoffrey Hinton &Oprah Winfrey \\
      Anne Hathaway & Jackie Chan &Renée Zellweger \\
      Audrey Hepburn & Jeff Bezos & Scarlett Johansson \\
    Barack Obama& Jet Li & Taylor Swift \\
Bill Gates& Joe Biden&  Thomas Edison \\
Bruce Li&Jürgen Schmidhuber&Vladimir Putin \\
Donald Trump&  Kamala Harris&Woody Allen \\
Dwayne Johnson& Marilyn Monroe &Yann LeCun \\
Elon Musk& Mark Zuckerberg & Yui Aragaki\\
    \bottomrule
    \end{tabular}
    \vspace{2mm}
    \caption{ID names used for evaluation.}
    \label{tab:evaluate_ids}
\end{table}
\begin{figure}[t]
    \centering
    \includegraphics[width=1\linewidth]{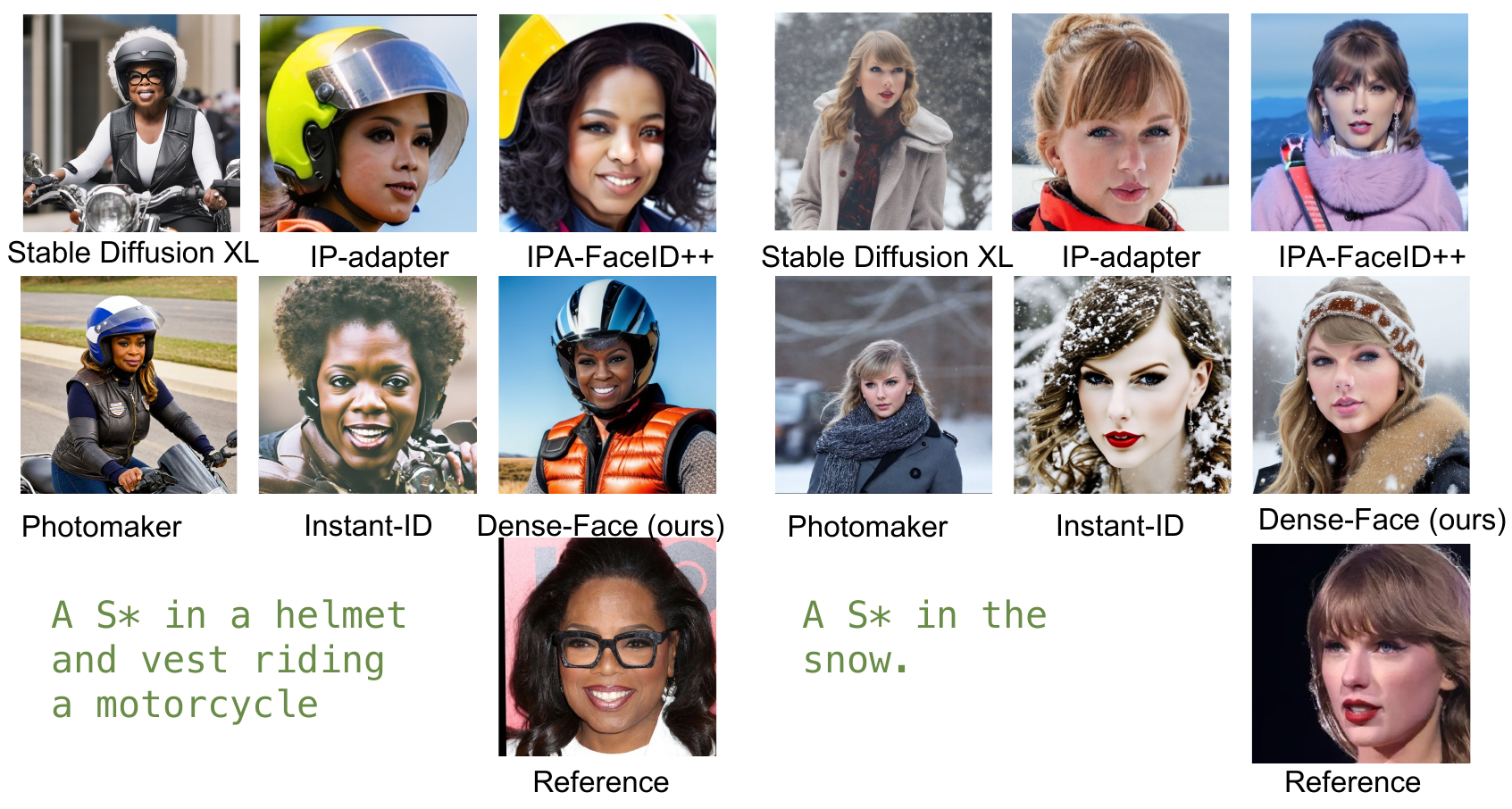}
    \includegraphics[width=1\linewidth]{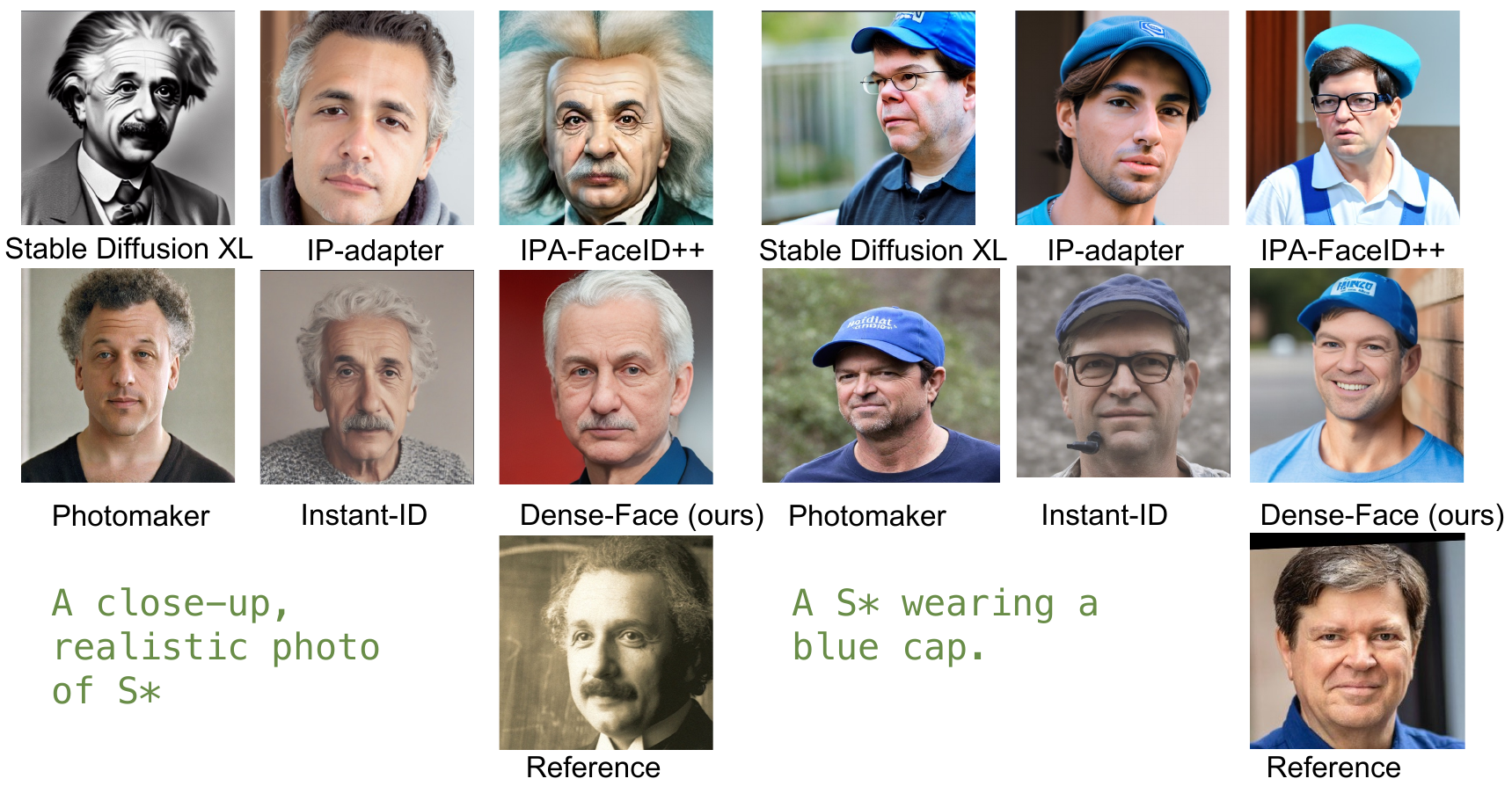}
    \caption{Additional comparisons among different personalized generation methods. Our proposed Dense-Face generates images with a consistent identity with the reference image, which can even be an old photo.}
    \label{fig_supp:comparison}
\end{figure}

\Paragraph{Implementation Details} We build up our Dense-Face based on pre-trained Stable Diffusion $2.1$, and the pose branch is based on the publicly available ControlNet. Specifically, we train the Dense-Face on T2I-Dense-Face with a batch size of $12$ and a learning rate of $5e-6$ on $8$ NVIDIA RTX A$6000$ GPUs for $3.5$ weeks. The $25$ steps DDIMScheduler with an improved noise schedule is used, and all images are cropped and resized into $512 \times 512$. 

\section{Additional Results}

\Paragraph{Personalized Geneartion}
Fig.~\ref{fig_supp:comparison} in the supplementary shows generation results from different personalized methods. Along with Fig.~\ref{fig:show_case_1} of the main paper, we can conclude that Dense-Face makes a competitive personalized generation method. Also, Fig.~\ref{fig_supp:personalization_v4} in the supplementary showcases more results of our text-controllable identity preservation generations. 

\Paragraph{Dense Annotation Prediction} We propose T2I-Dense-Face dataset in the Sec.~\ref{sec:dense_dataset} of the main paper. Specifically, we include different annotations and various parsed attributes of the given image, such as age, race, and gender. 
This is depicted in Fig.~\ref{fig_supp:dataset} of the supplementary. 
Our proposed dataset, which trains the Dense-Face, can be found on the project page. 
Dense-Face can produce different annotation predictions when generating the identity-preserved image, as depicted in Fig.~\ref{fig_supp:annotation} in the supplementary.

\Paragraph{Stylization} Following the previous work, we also apply our trained Dense-Face on various stylization generations, which is another important application of the personalized generation model. Results are shown in Fig.~\ref{fig_supp:style} of the supplementary.

\Paragraph{Face Swap} Aside from Tab.~\ref{fig:face_major} of the main paper, we use Fig.~\ref{fig_supp:face_swap} in the supplementary to show additional face swapping results. 

\begin{figure}[t]
    \centering
    \includegraphics[width=1.\linewidth]{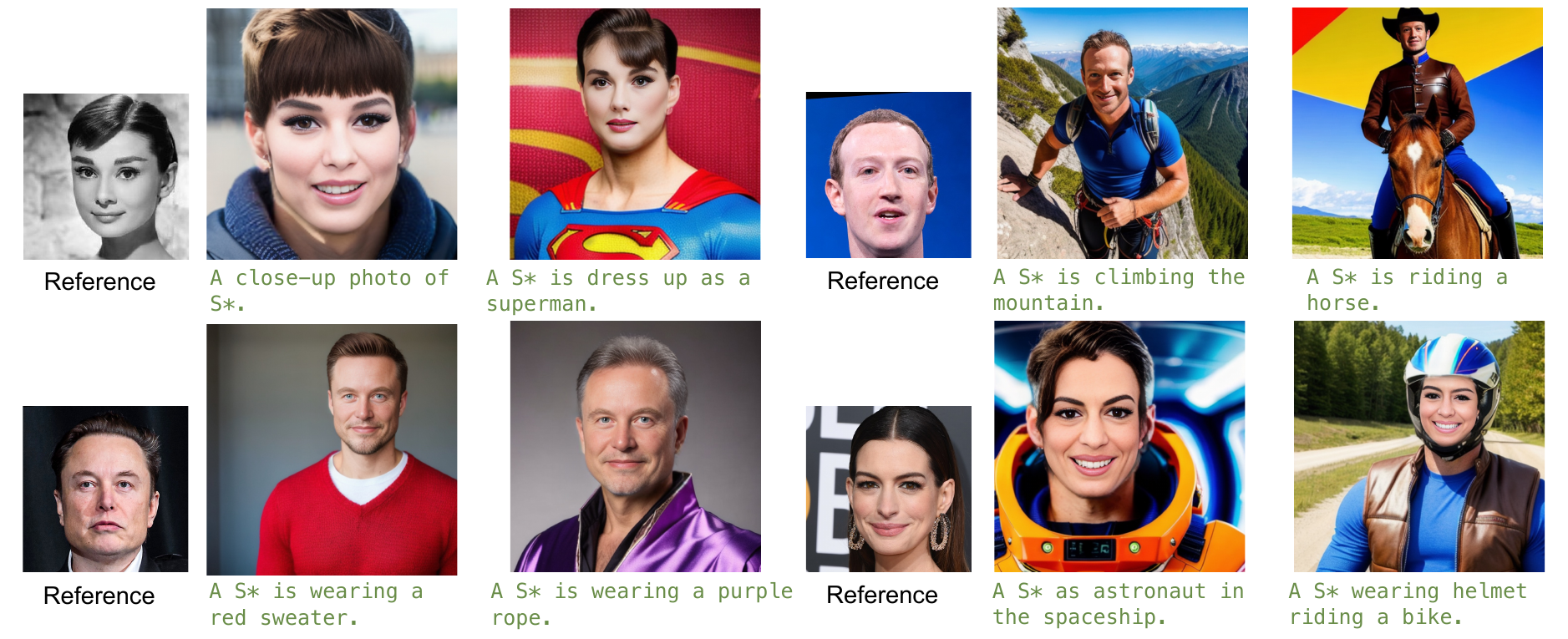}
    \includegraphics[width=1.\linewidth]{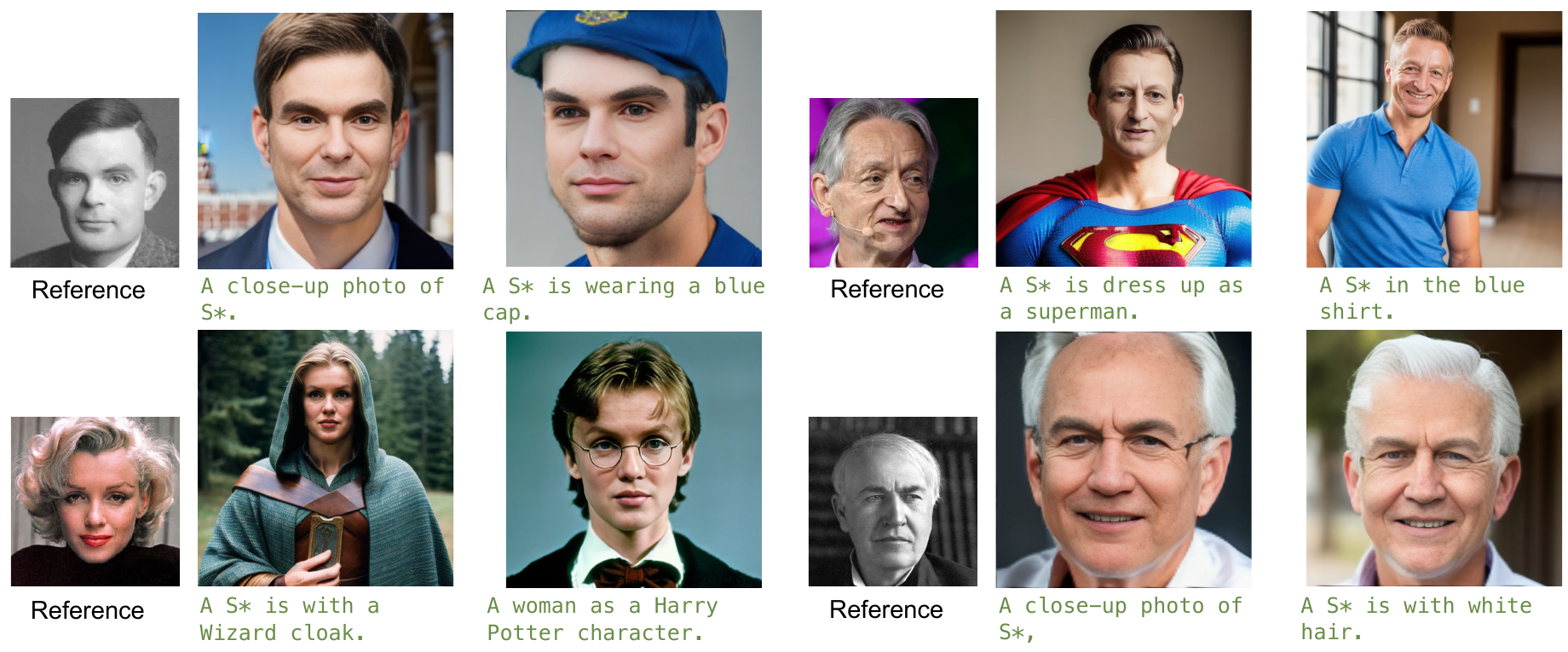}
    \vspace{-7mm}
    \caption{Dense-Face can place subjects in diverse contexts with changed attributes, such as hair color and clothes. \vspace{-6mm}}
    \label{fig_supp:personalization_v4}
\end{figure}
\begin{figure}[t]
    \centering
    \includegraphics[width=1\linewidth]{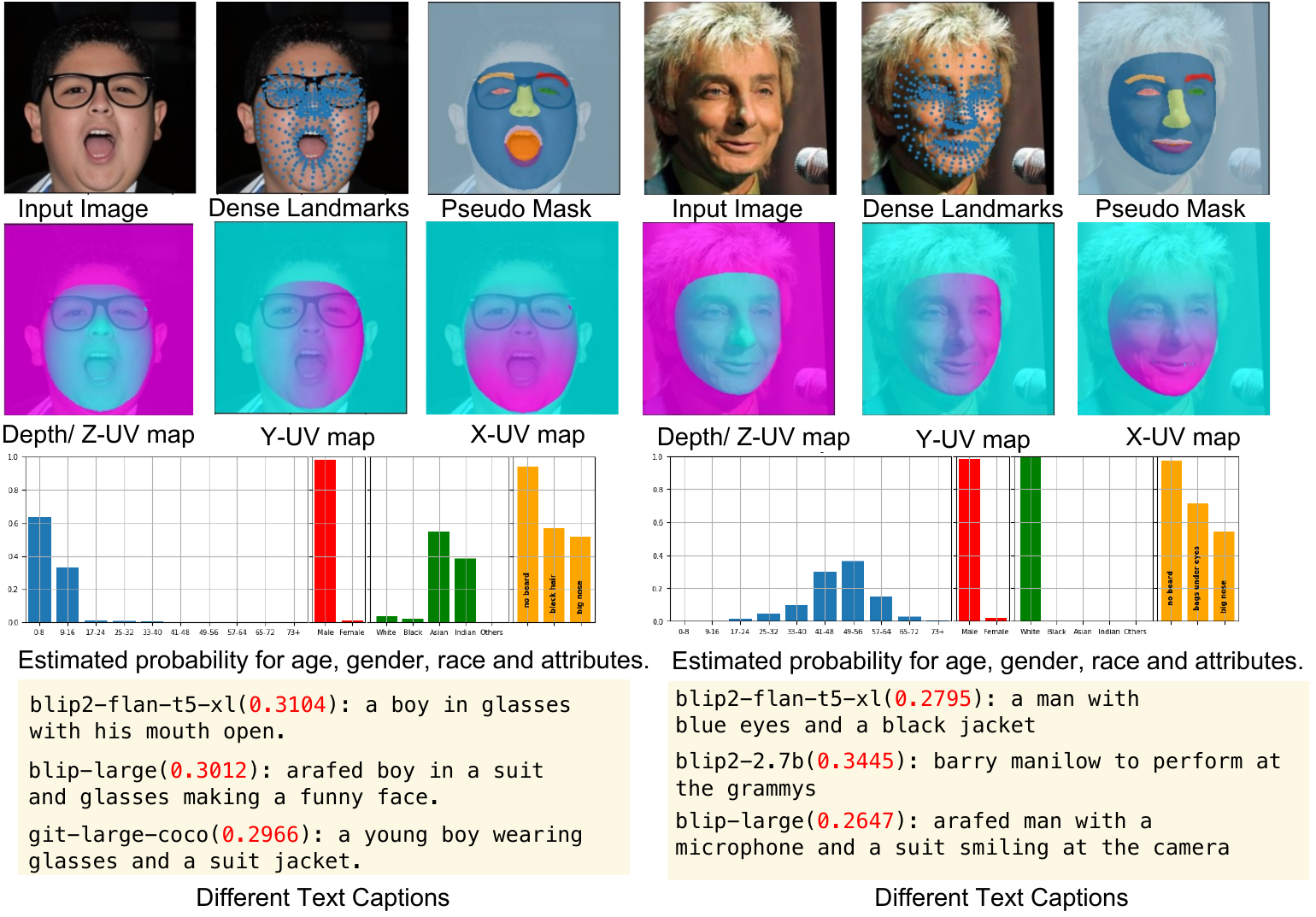}
    \caption{Two samples from the proposed T2I-Dense-Face.}
    \label{fig_supp:dataset}
\end{figure}
\begin{figure}[t]
    \centering
    \includegraphics[width=0.6\linewidth]{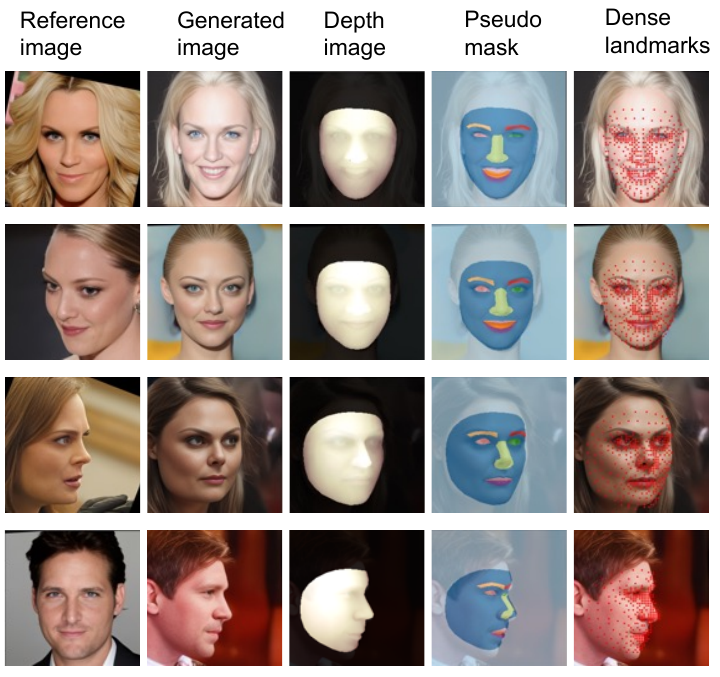}
    \caption{Additional visualizations on the dense annotation prediction. The proposed Dense-Face can generate high-fidelity identity-preserved images and corresponding annotations (\textit{e.g.}, depth image, pseudo mask, and landmarks). Generated images can be at large pose-views.}
    \label{fig_supp:annotation}
\end{figure}
\begin{figure}[t]
    \centering
    \includegraphics[width=1\linewidth]{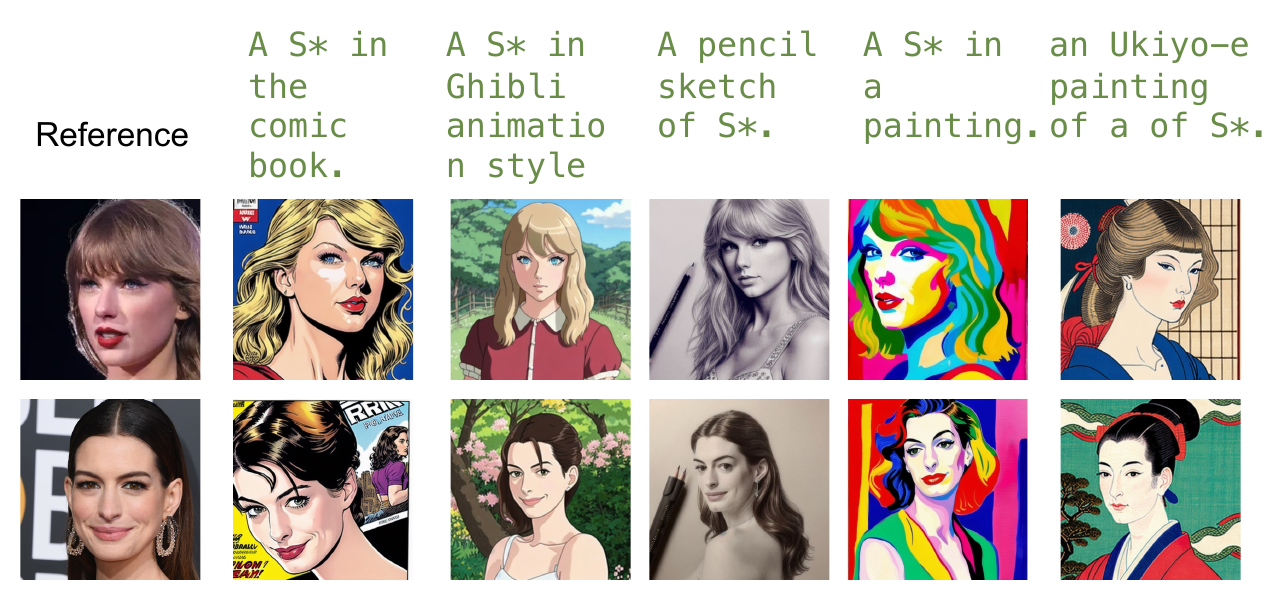}
    \includegraphics[width=1\linewidth]{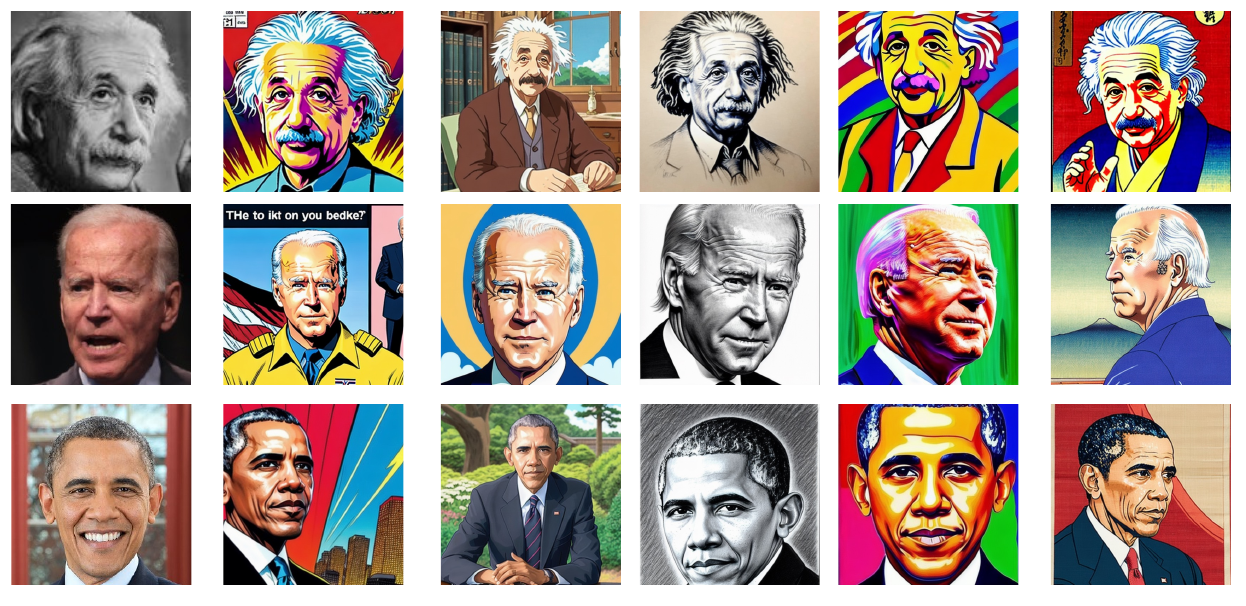}
    \caption{Additional visualizations on different subject stylizations.}
    \label{fig_supp:style}
\end{figure}
\begin{figure}[t]
    \centering
    \includegraphics[width=1.\linewidth]{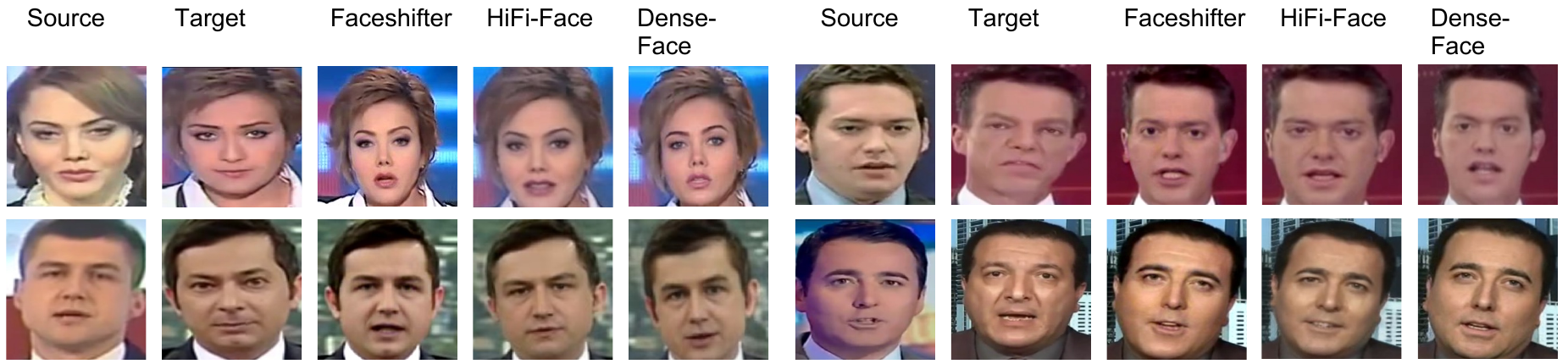}
    \caption{Additional Face Swapping results from the proposed method. Dense-Face achieves a comparable identity preservation to that of the previous work.}
    \label{fig_supp:face_swap}
\end{figure}
\clearpage
%
%
\bibliographystyle{splncs04}
\bibliography{main}
\end{document}